\documentclass[10pt,twocolumn,letterpaper]{article}

\usepackage{cvpr}      

\definecolor{iccvblue}{rgb}{0.21,0.49,0.74}
\usepackage[pagebackref,breaklinks,colorlinks,allcolors=iccvblue]{hyperref}

\usepackage[most]{tcolorbox}
\usepackage{xcolor}
\usepackage{graphicx}
\usepackage{booktabs}
\usepackage{makecell}
\usepackage{multirow}
\usepackage{adjustbox}

\definecolor{lightgray}{gray}{0.9}
\definecolor{boxblue}{RGB}{220, 230, 241}
\definecolor{boxyellow}{RGB}{255, 244, 204}
\definecolor{lightgreen}{RGB}{220, 250, 220}

\def\paperID{38} 
\def\confName{CVPR}
\def\confYear{2025}

\title{Synthetic Human Action Video Data Generation with Pose Transfer}

\author{
Vaclav Knapp\\
SSPS\\
{\tt\small Knapp.Va.2022@skola.ssps.cz}
\and
Matyas Bohacek\\
Stanford University\\
{\tt\small maty@stanford.edu}
}

\begin{document}
\twocolumn[{%
\renewcommand\twocolumn[1][]{#1}%
\maketitle

\centering
\begin{tabular}{p{0.03\textwidth}ccccc}
    \raisebox{0.3\height}{\rotatebox{90}{\shortstack{\small \textbf{Toyota} \\ \textbf{Smarthome}}}} &
    \includegraphics[width=0.16\linewidth]{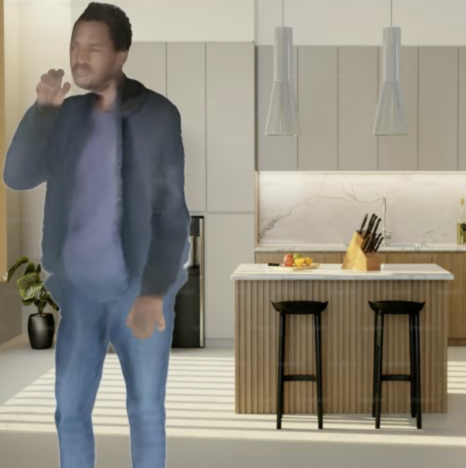} &
    \includegraphics[width=0.16\linewidth]{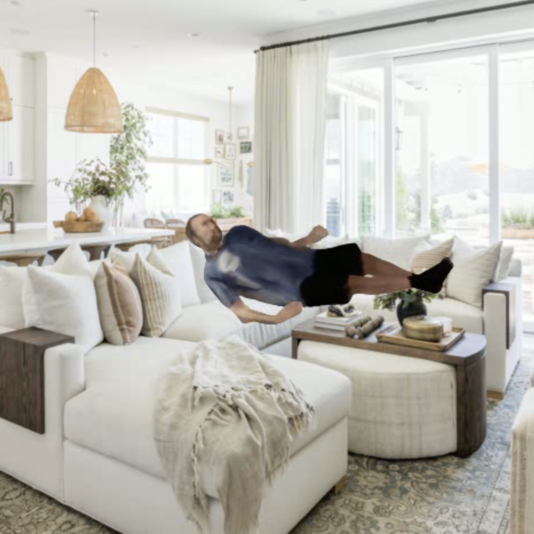} &
    \includegraphics[width=0.16\linewidth]{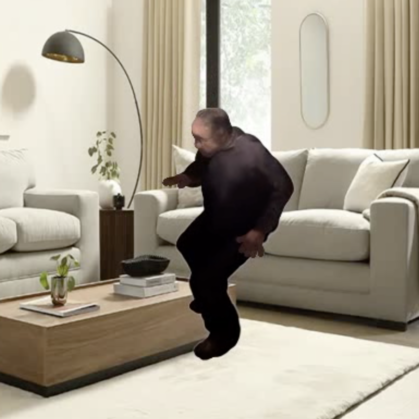} &
    \includegraphics[width=0.16\linewidth]{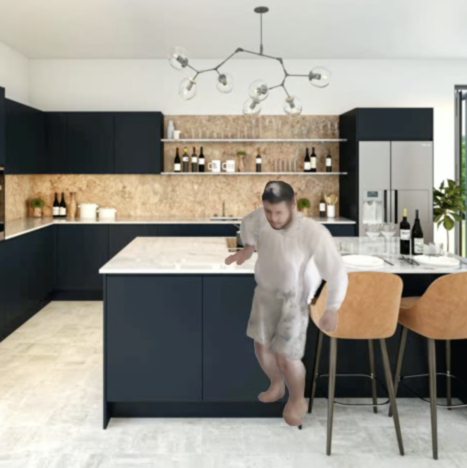} &
    \includegraphics[width=0.16\linewidth]{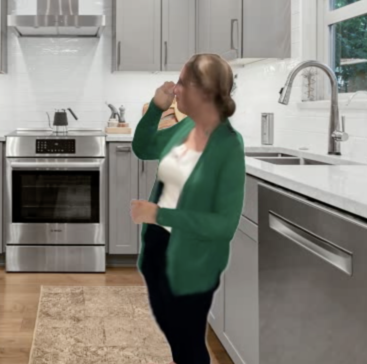} \\
    
    &\small Drink from bottle & 
    \small Lay down & 
    \small Sit down &
    \small Get up &
    \small Use phone \\ \\

    \raisebox{0.17\height}{\rotatebox{90}{\shortstack{\small \textbf{NTU RGB+D}}}} &
    \includegraphics[width=0.16\linewidth]{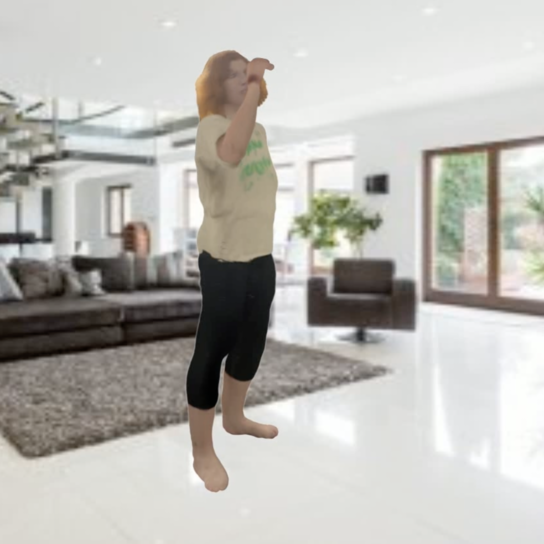} &
    \includegraphics[width=0.16\linewidth]{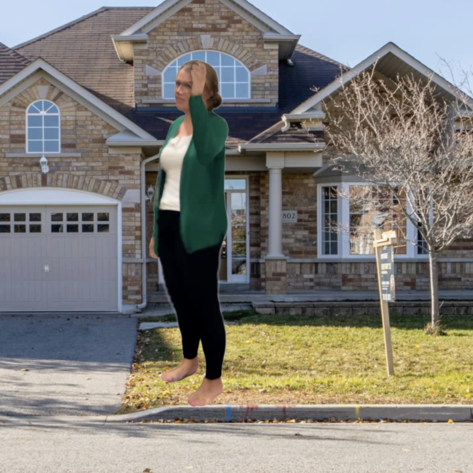} &
    \includegraphics[width=0.16\linewidth]{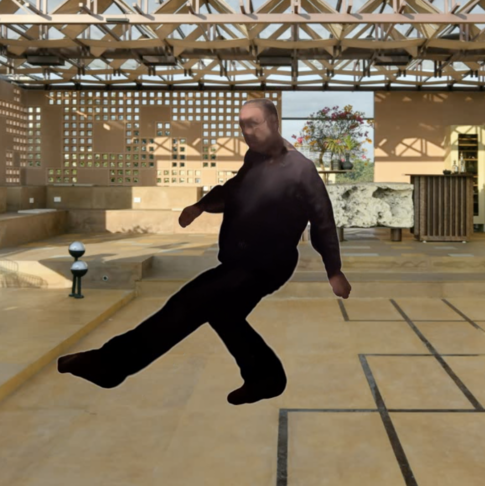} &
    \includegraphics[width=0.16\linewidth]{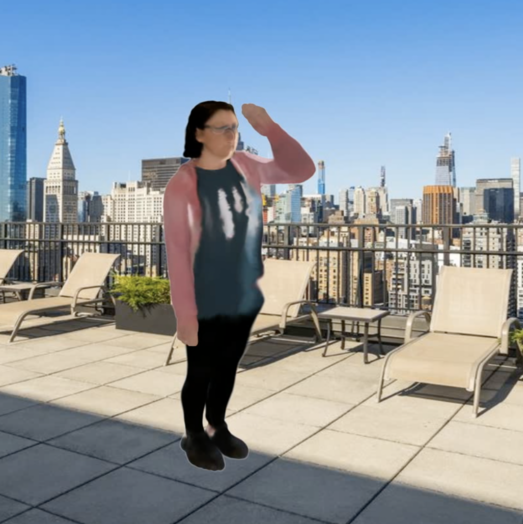} &
    \includegraphics[width=0.16\linewidth]{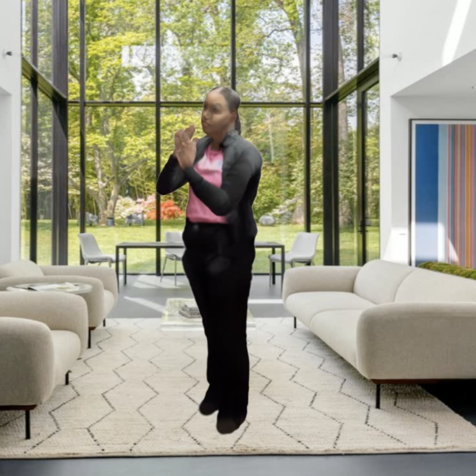} \\
    
    & \small Drink water & 
    \small Hand waving & 
    \small Kicking something &
    \small Salute &
    \small Put palms together \\
    
\end{tabular}

\captionof{figure}{Examples of synthetic video frames showing $12$ distinct human actions, generated using our method as extensions to the Toyota Smarthome (above) and the NTU RGB+D (below) datasets. The presented identities are from our \textit{RANDOM People} dataset. \vspace{1em}}

\label{fig:teaser}

}]

\begin{abstract}

In video understanding tasks, particularly those involving human motion, synthetic data generation often suffers from uncanny features, diminishing its effectiveness for training. Tasks such as sign language translation, gesture recognition, and human motion understanding in autonomous driving have thus been unable to exploit the full potential of synthetic data. This paper proposes a method for generating synthetic human action video data using pose transfer (specifically, controllable 3D Gaussian avatar models). We evaluate this method on the Toyota Smarthome and NTU RGB+D datasets and show that it improves performance in action recognition tasks. Moreover, we demonstrate that the method can effectively scale few-shot datasets, making up for groups underrepresented in the real training data and adding diverse backgrounds. We open-source the method along with \textit{RANDOM People}, a dataset with videos and avatars of novel human identities for pose transfer crowd-sourced from the internet.

\end{abstract}
    
\section{Introduction}
\label{sec:intro}

As large-scale training datasets proved essential for generalization in AI models~\cite{wang2021largescale}---be it for language, image, or video tasks---many internet-scraped datasets emerged. In domains where data at such a scale is unavailable, data augmentation and synthetic data stepped in to fill the gap~\cite{datacamp2024synthetic}. The recent advent of generative AI models has only accelerated this trend~\cite{xu2024generative}. However, in video understanding tasks, data augmentation offers only limited gains, and so far, approaches to whole-cloth synthetic data generation have fallen short of quality and overall usability~\cite{li2020data}. As a result, synthetic data is not widely used to train video classification or understanding models~\cite{jones2023synthetic}.

Much of the recent work in synthetic video data generation has focused on videos with human actions~\cite{zhang2022learn2augment}. This restricts the problem scope while still offering many use cases to downstream applications involving human action understanding and generation (e.g., sign language translation, gesture recognition, and human motion understanding in autonomous driving). In particular, two lines of work have prevailed in the literature over recent years: (1) approaches based on classic computer graphics, often using simulated rendering, and (2) generative AI models~\cite{liu2022promptonomyvit}. While the first approach preserves semantic attributes of the action (in that, the pose of protagonist in the video is fully controlled, and so the desired action is, too), it has not yet passed the uncanny valley~\cite{jones2023uncanny}. The second approach, on the other hand, can generate photorealistic human videos but is unstable in that it lacks full pose control and still contains physical errors as well as artifacts~\cite{xu2024photorealism}. Therefore, neither of these approaches has been widely adopted as a source of synthetic data for training video classification or understanding models.

Recent work has, however, focused on bridging these two approaches by leveraging the versatility and ability to generate photorealistic content that comes with generative AI while grounding it in a physical model of the human body that gives one full control over the resulting action~\cite{liu2022controlled,moon2024expressivewholebody3dgaussian}. While these models still have some issues (namely, high demands for reference identity videos and computational demands, as further discussed in Section~\ref{subsec:pose_transfer}), we found them to be effective enough to open the inquiry into using such models for synthetic video generation of human action for model training.

In particular, we devise a method that, given reference videos, reenacts the human actions shown in these reference videos using novel human identities in novel settings. To do so, we employ a modified ExAvatar~\cite{moon2024expressivewholebody3dgaussian} 3D Gaussian framework as our avatar animation backbone. We evaluate this method on action recognition using the Toyota Smarthome~\cite{das2019toyota} and NTU RGB+D~\cite{shahroudy2016ntu} datasets, improving the performance of two baseline models while proving particularly efficient for few-shot learning.

The primary contributions of this paper include:

\begin{enumerate}
    \item proposing a method for synthetic human action video data generation using pose transfer;
    \item open-sourcing this method with ExAvatar~\cite{moon2024expressivewholebody3dgaussian} as the avatar animation backbone;
    \item collecting and open-sourcing the \textit{RANDOM People} dataset with (a) novel human identity videos, (b) their avatars for pose reenactment, and (c) background images.
\end{enumerate}

\noindent The dataset, code, and additional resources are available at \url{synthetic-human-action.github.io}.

\section{Related Work}
\label{sec:related_work}

We open this section by reviewing the existing scholarship on synthetic data generation for action recognition model training. We continue with an overview of pose transfer methods for generating novel images and videos of people reenacting reference poses. We also enumerate the most prominent datasets for video action recognition. We close with a broader discussion of video classification models.
 
\subsection{Synthetic data for Action Recognition}

Early work on synthetic video data generation for action recognition model training exploited fully simulated environments powered by classic CGI methods. One of these methods is ElderSim~\cite{hwang2021eldersim}, specifically designed to generate videos of elderly individuals performing day-to-day tasks. Due to the uncanny features of the simulation, the resulting videos lacked photorealism and detail. Moreover, as these are fully simulated actions, they lack much of the imperfections of real-world human motion.

Another approach built around CGI simulation was employed in the creation of the Robot Control Gestures (RoCoG-v2) dataset~\cite{reddy2023synthetic}. This dataset comprises both real and synthetic videos of seven gesture classes for human-robot teaming. Similar to ElderSim, the method behind this dataset lacks options for adding external identities or matching real-world human actions, as it is powered by a CGI simulator.

Over time, methods based on classic CGI technologies have been extended and combined with AI techniques. This was the case with SURREACT~\cite{varol2021synthetic}, a model for generating synthetic videos from unseen viewpoints. While this approach has been shown to improve the robustness of action recognition models to diverse viewpoints, the method cannot generate completely new renditions of the actions. The target identities match those from the videos, but additional external identities are not supported.

More recently, BEDLAM~\cite{black2023bedlam} showed that a purely synthetic dataset built using SMPL-X~\cite{pavlakos2019expressive} can be used to achieve state-of-the-art performance on human pose estimation. SynthAct~\cite{schneider2024synthact} further showed how Unity can be leveraged to create synthetic human action video data suitable for robotics applications.

While these recent advancements highlight the potential of synthetic data for certain computer vision tasks, most existing methods that generate synthetic data for action recognition still struggle with photorealism and generalization to new target identities. Although they have been shown to enhance human action video classifiers in some scenarios (e.g., handling novel viewpoints), achieving this often requires substantial adaptation to the task at hand. As described Section~\ref{sec:methods}, our method overcomes many of these shortcomings while utilizing the advantages of explicit human models, which ensure full control of the generated human motion. Furthermore, it is versatile and can be applied to many contexts without modifications.

\subsection{Pose Transfer}
\label{subsec:pose_transfer}

The task of pose transfer entails generating images or videos of a given human identity (target) in new poses (source). Formulations of this task differ in the input (target identity and source pose) and output modality.

Earlier approaches focused only on generating static images. The first successful works in this domain broke ground around 2017 when a two-stage neural network for generating images of people conditioned on pose was introduced~\cite{ma2017pose}. From then on, the task has been approached by emerging neural network architectures, as seen in other areas of computer vision: most prominently, GANs~\cite{siarohin2018deformable,men2020controllable,liu2019liquid} and attention-equipped methods~\cite{zhu2019progressive,roy2022multi}.

Recent approaches focused on generating videos. These include Magic Animate \cite{xu2024magicanimate}, Animate Anyone \cite{hu2024animate}, Champ \cite{zhu2024champ}, and ExAvatar \cite{moon2024expressivewholebody3dgaussian}, which we review in more detail.

Magic Animate extracts a DensePose~\cite{guler2018densepose} representation of each frame in the source video. The target identity is provided as a single image. To generate novel videos of the target identity in the source poses, Magic Animate employs a video diffusion model with an appearance encoder.

Animate Anyone extracts body keypoints of each frame in the source video. The target identity is provided as a single image. To generate novel videos of the target identity in the source poses, Animate Anyone trains a 3D U-Net model for video denoising and incorporates an additional reference network to extract features from the reference image.

Champ extracts an SMPL model~\cite{loper2023smpl} (a parametric 3D pose representation) from each frame in the source video. The target identity is provided as a single image. To generate novel videos of the target identity in the source poses, Champ leverages SMPL as a latent representation within a custom-trained diffusion model.

ExAvatar extracts an expressive whole-body 3D Gaussian representation of each frame in the source video, which is obtained through an ensemble of other body representations, from body keypoints to SMPL-X~\cite{pavlakos2019expressivebodycapture3d} to expressive head models. The target identity is provided as a video of the person performing a specific set of actions; from this video, the same ensemble of body representations is extracted. Once the avatar and source pose representations are extracted, 3D Gaussian Splatting allows ExAvatar to animate the target identity according to the source poses. In a recent study, human evaluators found ExAvatar to produce human motion that is more consistent with the reference and overall more coherent than its diffusion-based counterparts Magic Animate and Animate Anyone~\cite{knapp2025can}.

On the one hand, Magic Animate, Animate Anyone, and Champ are easier to scale because they only analyze a single format of input representations and require only a single image of the target identity, unlike ExAvatar, which analyzes an ensemble of features and requires a specific kind of video of the target. However, constructing 3D Gaussian avatars gives ExAvatar full, guaranteed control of the pose, while Magic Animate, Animate Anyone, and Champ rely on generative approaches like diffusion where the control is not guaranteed. Moreover, these diffusion-based methods often suffer from artifacts, hallucinations, and other deficiencies.

\subsection{Video Datasets for Action Recognition}

There are multiple popular video datasets for action recognition in the literature. We highlight four most prominent ones, in chronological order of their release.

\textbf{HMDB51}~\cite{kuehne2011hmdb} contains a total of $\text{6,766}$ videos across $\text{51}$ classes, collected from YouTube. The actions in this dataset are diverse, including sports, musical instrument playing, and close-up interactions with objects.

\textbf{UCF101}~\cite{Soomro2012UCF101AD} contains a total of $\text{13,320}$ videos across $\text{101}$ classes, from various online sources. Similar to HMDB51, UCF101 also includes a wide range of actions.

\textbf{NTU RGB+D}~\cite{shahroudy2016ntu} contains a total of $\text{56,880}$ videos across $\text{60}$ classes recorded by the dataset authors with consenting protagonists. The protagonists are shown performing a wide range of regular daily actions (e.g., walking and drinking).

\textbf{SSV2}~\cite{goyal2017something} contains a total of $\text{220,847}$ videos across $\text{174}$ classes. The dataset contains videos of many fine-grained interactions between humans and objects (e.g., pushing, pulling, and sliding).

\textbf{Toyota Smarthome}~\cite{das2019toyota} contains a total of $\text{16,115}$ videos across $\text{31}$ classes, with $\text{18}$ unique participants. This dataset shows mostly senior citizens perform daily activities inside of their homes. This dataset is distinct for its notable intra-class variation and class imbalance.

In our experiments, we chose to use Toyota Smarthome and NTU RGB+D because they include only consenting participants and do not rely on object interactions.

\subsection{Video Classification Models}

Popular baseline models for video classification include I3D~\cite{carreira2017quo}, ResNet adapted for video processing~\cite{hara2018can}, SlowFast~\cite{feichtenhofer2019slowfast}, and, more recently, architectures derived from the Vision Transformer (ViT)~\cite{dosovitskiy2020image, arnab2021vivit, bertasius2021space, tong2022videomae}. Many ViT-based models are pre-trained on the Kinetics dataset~\cite{carreira2017quo} for action recognition, facilitating downstream training on datasets such as Toyota Smarthome and NTU RGB+D. Given this, we opt for older baselines---ResNet and SlowFast---and train them from randomly initialized weights.




\begin{figure*}[t]
    \centering
    \includegraphics[width=\linewidth]{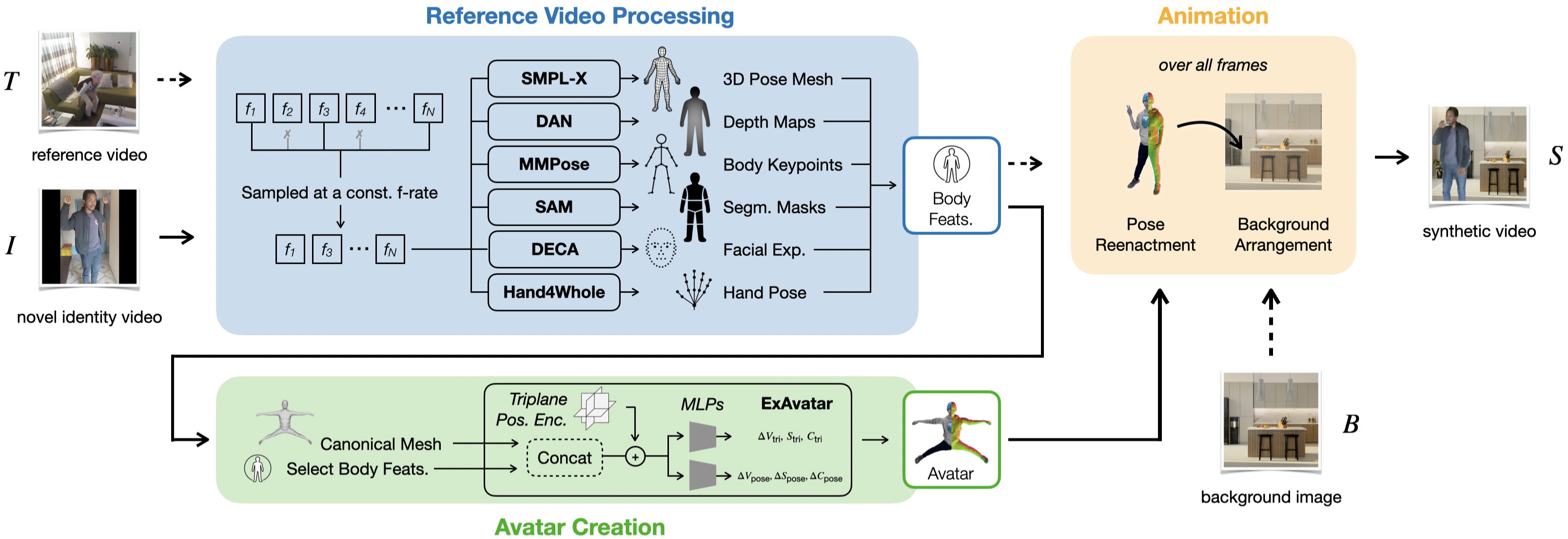}
    
    \caption{Overview of our method for synthetic human action video data generation. Avatar example taken from~\cite{moon2024expressivewholebody3dgaussian}. See Section~\ref{sec:methods}.}
    \label{fig:overview}
    
\end{figure*}

\section{Methods}
\label{sec:methods}

In this section, we describe our method for synthetic data generation of human action videos. The method extends existing datasets by using real videos as reference human action that is reenacted by novel identities in new settings. \\

An overview diagram of the method is shown in Figure~\ref{fig:overview}. At the input, the method receives $T$, a set of $n_T$ real reference videos with human actions, as well as $I$, a set of $n_I$ videos with novel human identities for action reenactment. The videos in $I$ show people performing a sequence of actions that was found to be effective for avatar generation in~\cite{moon2024expressivewholebody3dgaussian}. Optionally, the method also receives a set of background scene images $B$. It then generates new videos in which identities in $I$ are animated to reenact the same actions as shown in the real reference videos $T$, optionally with varying background settings $B$. The method consists of three main stages: (1) avatar creation, (2) reference video preparation, and (3) animation.

\subsection{Avatar Creation}
\label{subsec:avatar_creation}

For each novel human identity $I_i$ in $I_1 \dots I_{n_I}$, an expressive whole-body avatar $A_i$ is made using ExAvatar~\cite{moon2024expressivewholebody3dgaussian}, which we use as our avatar animation backbone. This yields fully controllable 3D Gaussian avatars.

To that end, the video $I_i$ is normalized to a constant length and frame rate. Select frames within the normalized video are then used to extract the following features (the frames selection follows ExAvatar's original implementation~\cite{moon2024expressivewholebody3dgaussian}):

\begin{enumerate}
    \item 3D pose meshes (SMPL-X~\cite{pavlakos2019expressivebodycapture3d});
    \item depth maps (DepthAnythingv2~\cite{yang2024depthv2});
    \item body keypoints ($133 \cdot 3$, MMPose~\cite{mmpose2020});
    \item identity segmentation masks (Segment Anything~\cite{kirillov2023segany});
    \item facial expressions (DECA~\cite{feng2021learninganimatabledetailed3d});
    \item hand poses (Hand4Whole~\cite{moon2022accurate3dhandpose}).
\end{enumerate}

With these features extracted, we train $A_i$, an animatable 3D Gaussian avatar that preserves the unique identity characteristics of the novel human identity in $I_i$. In doing so, we follow the implementation of ExAvatar~\cite{moon2024expressivewholebody3dgaussian} with default parameters unless specified above.

\subsection{Reference Video Preparation}

Each reference video $T_i$ in $T_0 \dots T_{n_T}$ is normalized to a constant length and frame rate. Select frames within the normalized video are then used to extract the same set of features as used for avatar creation (described above in Section~\ref{subsec:avatar_creation}) with the exception of identity segmentation masks and depth maps, which are not extracted. These features are packaged into $F_i$.

\subsection{Animation}

With the set of novel identity avatars $A$ and the set of reference video features $F$ representing training videos $T$, we proceed with the animation stage.

\textbf{White-Background Videos.} For each training video $T_i$ in $T_1 \dots T_{n_T}$, represented by its features $F_i$, and each novel identity avatar $A_j$ in $A_1 \dots A_{n_A}$, we generate a synthetic video $S_{(i,j)}$. This results in $ n_A $ synthetic videos per training video, totaling $ n_T \cdot n_A $ synthesized videos. These intermediate videos will be superimposed onto image backgrounds in the next step.

\textbf{Image-Background Videos.} As a final step, image backgrounds are added to the white-background videos. For each training video $ T_i $ in $ T_1 \dots T_{n_T} $, represented by its features $ F_i $, and each novel identity avatar $ A_j $ in $ A_1 \dots A_{n_A} $, we randomly select $ g $ image backgrounds from a background pool $ B $, denoted as $ B_k $. For each combination of $ T_i $, $ A_j $, and $ B_k $, we synthesize the video $ S_{(i,j,k)} $. This results in $ g \cdot n_A $ synthetic videos per training video, for a total of $ n_T \cdot g \cdot n_A $ synthesized videos.
\begin{figure*}[t]
    \centering
    \includegraphics[width=0.16\textwidth]{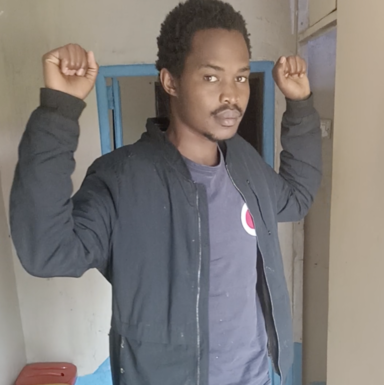}
    \hspace{0.01\textwidth}
    \includegraphics[width=0.16\textwidth]{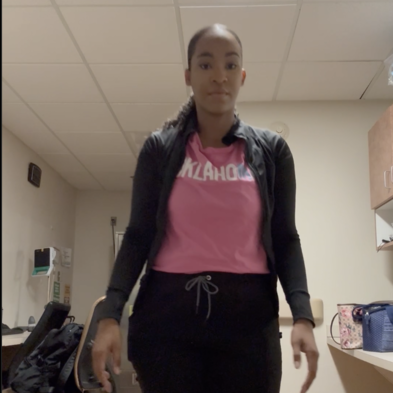}
    \hspace{0.01\textwidth}
    \includegraphics[width=0.16\textwidth]{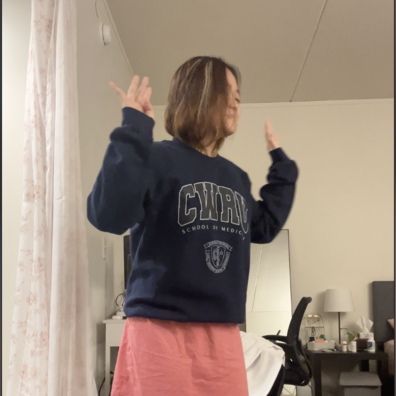}
    \hspace{0.01\textwidth}
    \includegraphics[width=0.16\textwidth]{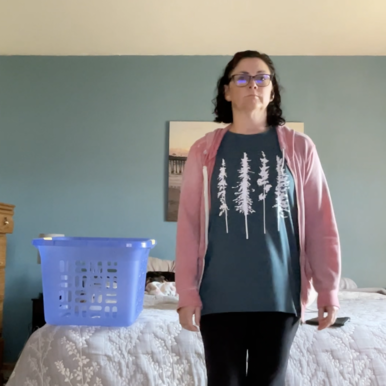}
    \hspace{0.01\textwidth}
    \includegraphics[width=0.16\textwidth]{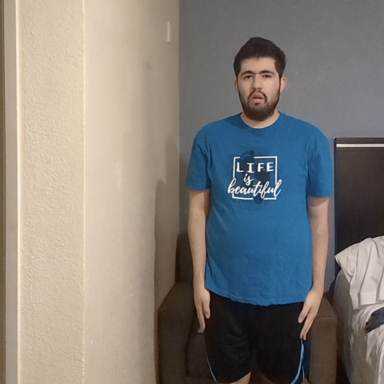}
    
    \caption{Representative frames from the novel human identity videos $I$ in the \textit{RANDOM People} dataset.}
    \label{fig:novel_human_identity_videos_examples}
\end{figure*}

\begin{figure*}[t]
    \centering
    \includegraphics[width=0.16\textwidth]{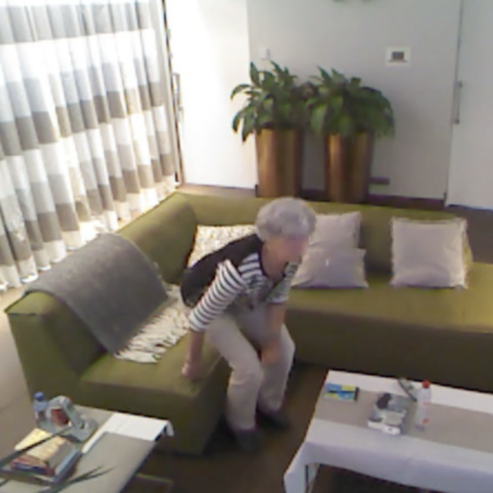}
    \hspace{0.01\textwidth}
    \includegraphics[width=0.16\textwidth]{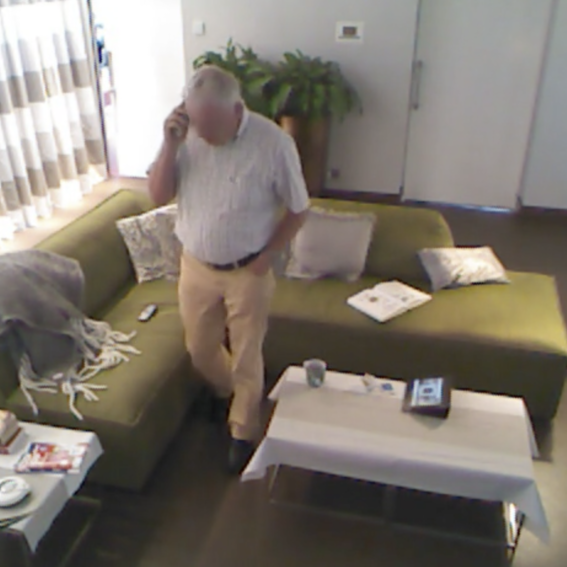}
    \hspace{0.01\textwidth}
    \includegraphics[width=0.16\textwidth]{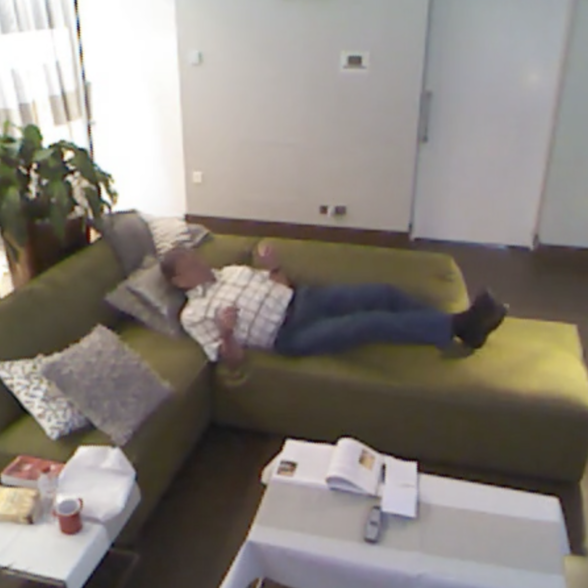}
    \hspace{0.01\textwidth}
    \includegraphics[width=0.16\textwidth]{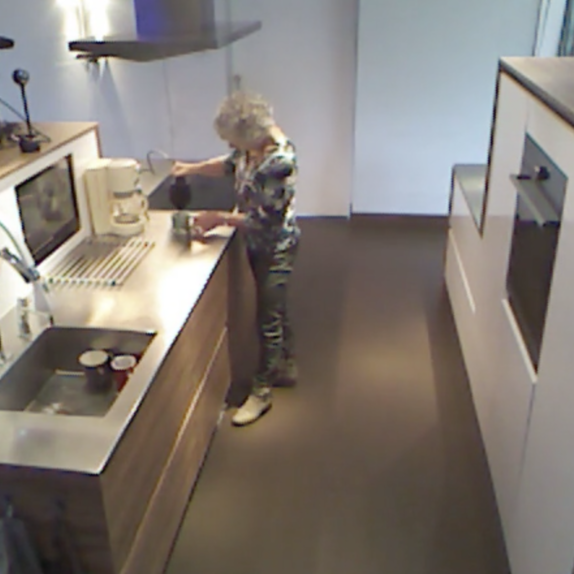}
    \hspace{0.01\textwidth}
    \includegraphics[width=0.16\textwidth]{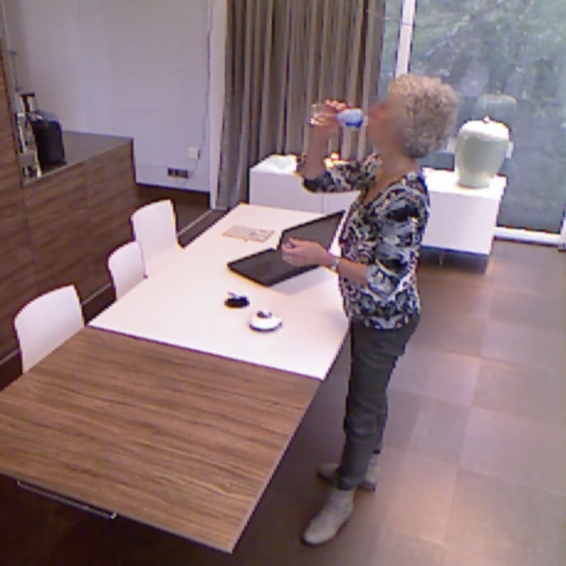}
    \caption{Representative frames from the reference videos $T$ in the \textit{RANDOM People} dataset. These videos, sourced from the Toyota Smarthome dataset, were manually selected following suitability criteria in Section~\ref{subsec:data__reference_human_action_videos}.}
    \label{fig:reference_human_action_videos_examples}
\end{figure*}

\begin{figure*}
    \centering
    \includegraphics[width=0.16\textwidth]{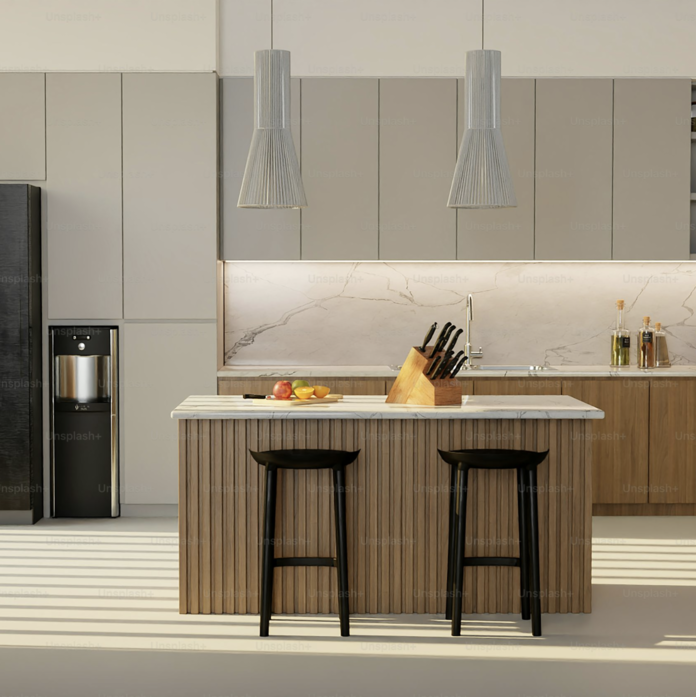}
    \hspace{0.01\textwidth}
    \includegraphics[width=0.16\textwidth]{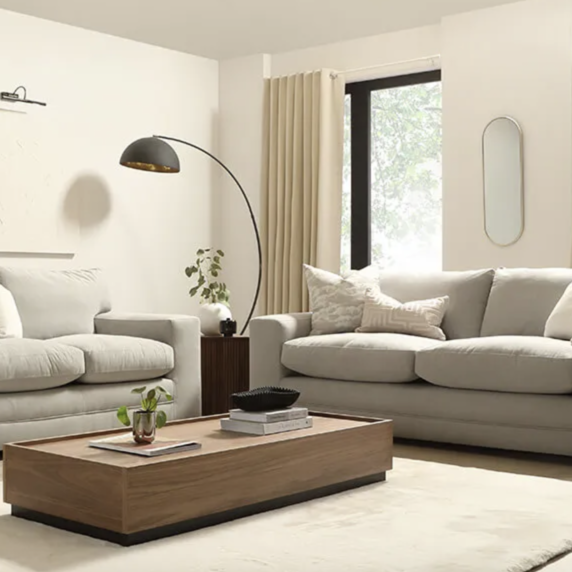}
    \hspace{0.01\textwidth}
    \includegraphics[width=0.16\textwidth]{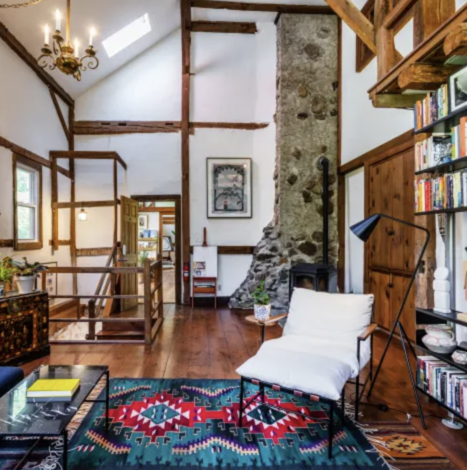}
    \hspace{0.01\textwidth}
    \includegraphics[width=0.16\textwidth]{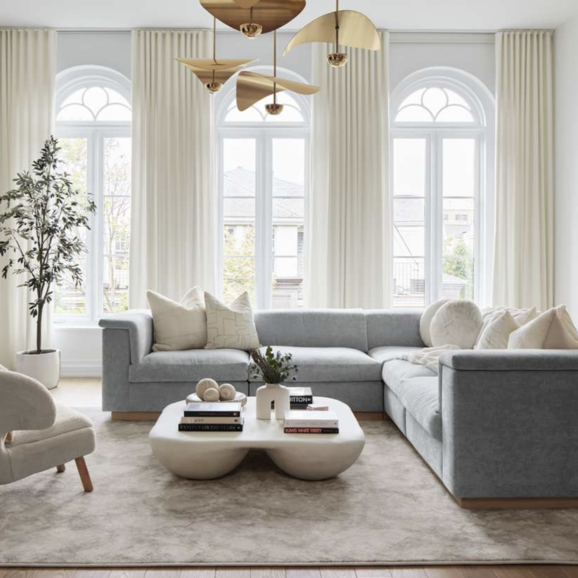}
    \hspace{0.01\textwidth}
    \includegraphics[width=0.16\textwidth]{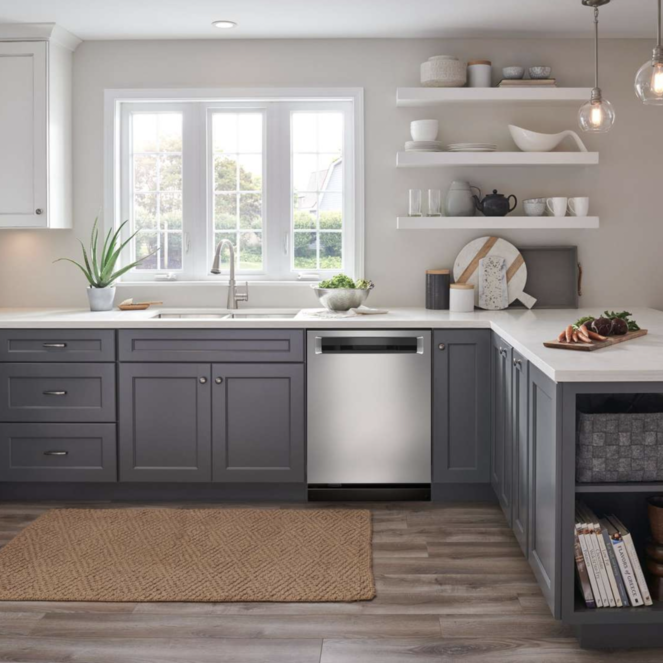}
    
    \caption{Representative background images from $B$ in the \textit{RANDOM People} dataset.}
    \label{fig:background_imgs_examples}
\end{figure*}

\section{Data}
\label{sec:data}

In this section, we describe the data used in our experiments. We first detail the data collection process of novel human identity videos $I$. We then describe the selection of reference videos $T$. Finally, we cover the selection of background scene images $B$.

\subsection{Novel Human Identity Videos}
\label{subsec:data__novel_human_identity_videos}

Recordings of $188$ participants were crowd-sourced through Prolific. These participants were informed about the intended use of their recordings and asked to consent to their video---as well as a 3D Gaussian model and other derived artifacts---made publicly available for research purposes prior to entering our interface. The average completion time was $8$ minutes, and participants were compensated at a rate of above $8$ USD. This data collection was consulted with the IRB office at Stanford University, which concluded that an IRB review is not required.

The participants were instructed to record themselves performing three slow $360\textdegree$ rotations, with a mix of raised and lowered hands. Representative examples of frames from these videos are shown in Figure~\ref{fig:novel_human_identity_videos_examples}. See Appendix~\ref{app:participant_instructions} for full recording instructions and acceptance criteria. Based on these criteria, we filtered the $188$ collected videos to a set of $100$ videos that met our criteria, used as novel human identity videos $I$.

The participants were chosen from a stratified sample of residents of the United States. Based on self-reported demographics, the dataset (after filtering) contains $41$ people who identity as male and $59$ people who identify as female. $11$ participants are Asian, $10$ participants are Black, $11$ participants are of mixed race, $66$ participants are white, and $2$ participants reported a different race. The median age is $38$, with a minimum age of $19$ and a maximum age of $64$. The average response time was $10$ minutes.


\subsection{Reference Human Action Videos}
\label{subsec:data__reference_human_action_videos}

The following procedure was performed with each evaluated dataset (Toyota Smarthome~\cite{Dai_2022_PAMI} and NTU RGB+D~\cite{shahroudy2016ntu}) individually. To curate a set of reference videos $T$, we manually selected a subset of $16$ human action classes from the dataset (see Appendix~\ref{app:action_classes} for the complete list). The selected classes met criteria outlined in Appendix~\ref{app:action_classes}. For each class, we then manually selected $5$ videos, yielding a total of $n_T = 16 \cdot 5 = 80$ videos. Representative examples of frames from these videos are shown in Figure~\ref{fig:reference_human_action_videos_examples}. 

\subsection{Background Images}

The pool of background images $B$ was created to match typical environments for actions present in each of the evaluated datasets. In total, $n_B = 20$ images were scraped from the internet for each evaluated dataset. Representative examples of background images in $B$ are shown in Figure~\ref{fig:background_imgs_examples}.

\subsection{Synthetic Human Action Videos}

Using $n_I = 15$ novel human identity videos $I$, $n_T = 80$ reference human action videos $T$, and $n_B = 40$ background image pool, sampled at $g = 3$ per video, we applied our synthetic data generation method to produce $S$: a set of $n_T \cdot g \cdot n_A = 3,600$ image-background synthetic videos for each evaluated dataset. Representative frames from these videos are shown in Figure~\ref{fig:teaser}.

Upon manual review, we found that most synthetic videos were consistent with the source videos in terms of pose and scene placement, as illustrated in Figure~\ref{fig:consistency_frame_grid_consistent} (Appendix~\ref{app:qualitative_eval}). However, in some cases, the generated videos deviated from the source poses and scene alignment, as shown in Figure~\ref{fig:consistency_frame_grid_inconsistent} (Appendix~\ref{app:qualitative_eval}).

We used the smaller \textit{RANDOM People 15} subset due to computational constraints. See Appendix~\ref{app:infra_and_runtime_considerations} for further details on compute considerations.

\subsection{Open-sourcing}

We call our dataset \textit{RANDOM People} and open-source it for research purposes at \url{synthetic-human-action.github.io}. This data release includes the synthesized videos $S$, the underlying novel human identity videos $I$ along with their avatars $A$, and the scene background images $B$. We designate two subsets of the dataset: \textit{RANDOM People 15} with $15$ identities, intended for small-scale experiments, and \textit{RANDOM People 100} with $100$ identities, intended for large-scale experiments. The dataset is made available under the CC BY-NC 4.0 license\footnote{This license allows use, adaptation, and sharing of the dataset provided the use is non-commercial and appropriate credit is given. See \url{https://creativecommons.org/licenses/by-nc/4.0/deed.en} for details.}.


\begin{table}[t]
    \centering
    \begin{tabular}{c|p{1cm}p{1cm}|cc}
        \toprule
        & \rotatebox{45}{Original}
        & \rotatebox{45}{Synthetic}
        & Toyota & NTU RGB+D \\
        \midrule
        \multirow{2}{*}{\textbf{ResNet}} 
        & \checkmark & & 20.46 & 8.66 \\
        & \checkmark & \checkmark & 51.15 & 42.98 \\
        \midrule
        \multirow{2}{*}{\textbf{SlowFast}} 
        & \checkmark & & 38.35 & 26.75 \\
        & \checkmark & \checkmark & 55.64 & 36.29 \\
        \bottomrule
    \end{tabular}
    \caption{Testing accuracy of ResNet and SlowFast on the Toyota and NTU RGB+D subsets trained in two configurations: with only the original data and both the original and synthetic data.}
    \label{tab:results}
\end{table}

\section{Experiments}
\label{sec:experiments}

We conducted three sets of experiments---baseline, one-shot, and few-shot---on both Toyota Smarthome and NTU RGB+D.

\subsection{Baseline Experiments}

We performed the baseline experiments on two standard video classification architectures: ResNet adapted for video processing~\cite{hara2018can} and SlowFast~\cite{feichtenhofer2019slowfast}. In one case, the model was trained only on the original data from the respective dataset; in the other, it was trained on both the original data and our synthetic data. The training set consisted of $n_{\text{real}} = n_{\text{background}} = 225$ videos per class\footnote{We capped the number of samples across data sources at $225$ (i.e., $n_{\text{real}} = n_{\text{background}} = 225$), despite having more synthetic samples, to match the minimum per-class repetition count in the Toyota Smarthome dataset. This ensured that the ablation compared the contributions of each component rather than reflecting imbalanced data source ratios.}. The testing set, composed only of original videos, included $n_{test} = 50$ videos per class.

The model was trained for $5$ epochs with a learning rate of $1\times10^{-4}$ and a batch size of $4$. The input frames were resized to $224\times224$. The ResNet model had a depth of $50$, with the number of frames set to $16$. For SlowFast, the number of frames was set to $32$. If not otherwise specified, default hyperparameter values from PyTorchVideo~\cite{fan2021pytorchvideo} were used (a snapshot is included in the open-source release).


\subsection{One-shot Experiments}

We further evaluated the effectiveness of our synthetic data generation method in a one-shot learning scenario on ResNet. With $n_{\text{real}} =1$, we trained five models while increasing the sample of synthetic videos from $n_{\text{background}}=0$ to $n_{\text{background}}=200$ by $50$ videos at a time. The testing set was the same as in the baseline configuration; the one-shot example was selected at random.

\subsection{Few-shot Experiments}

Finally, we evaluated the effectiveness of our synthetic data generation method in a few-shot learning scenario. The setup is identical as in the previous one-shot experimental setup, except $n_{\text{real}} =5$. These few-shot samples, too, are selected at random.


\section{Results}
\label{sec:results}

In this section, we report the results of our experiments, implemented using the PyTorch~\cite{paszke2019pytorch} and PyTorchVideo~\cite{fan2021pytorchvideo} libraries. The source code is fully open-sourced at \url{synthetic-human-action.github.io}.

\subsection{Baseline Experiments}

Table~\ref{tab:results} presents the classification accuracy for each model when trained with only original data and with original plus synthetic data.

When trained solely on the original data, ResNet achieved an accuracy of $20\%$ on Toyota and $9\%$ on NTU RGB+D.\footnote{We investigated this performance outside of the baseline methodology to understand why it was below chance ($6.25\%$ for this $16$-class task). We found that the model struggled to converge, even with a hyperparameter search. While the model occasionally exceeded $10\%$ accuracy on the validation set after some epochs, it consistently regressed below this threshold.} When synthetic data was added, the accuracy increased to $51\%$ and $43\%$, respectively.

Similarly, when trained solely on the original data, SlowFast achieved an accuracy of $38\%$ on Toyota and $27\%$ on NTU RGB+D. With the addition of synthetic data, performance improved to $56\%$ and $36\%$, respectively.

These results demonstrate that incorporating our synthetic data enhances the performance of video classification models in action recognition.

\begin{figure}
    \centering
    \includegraphics[width=\linewidth]{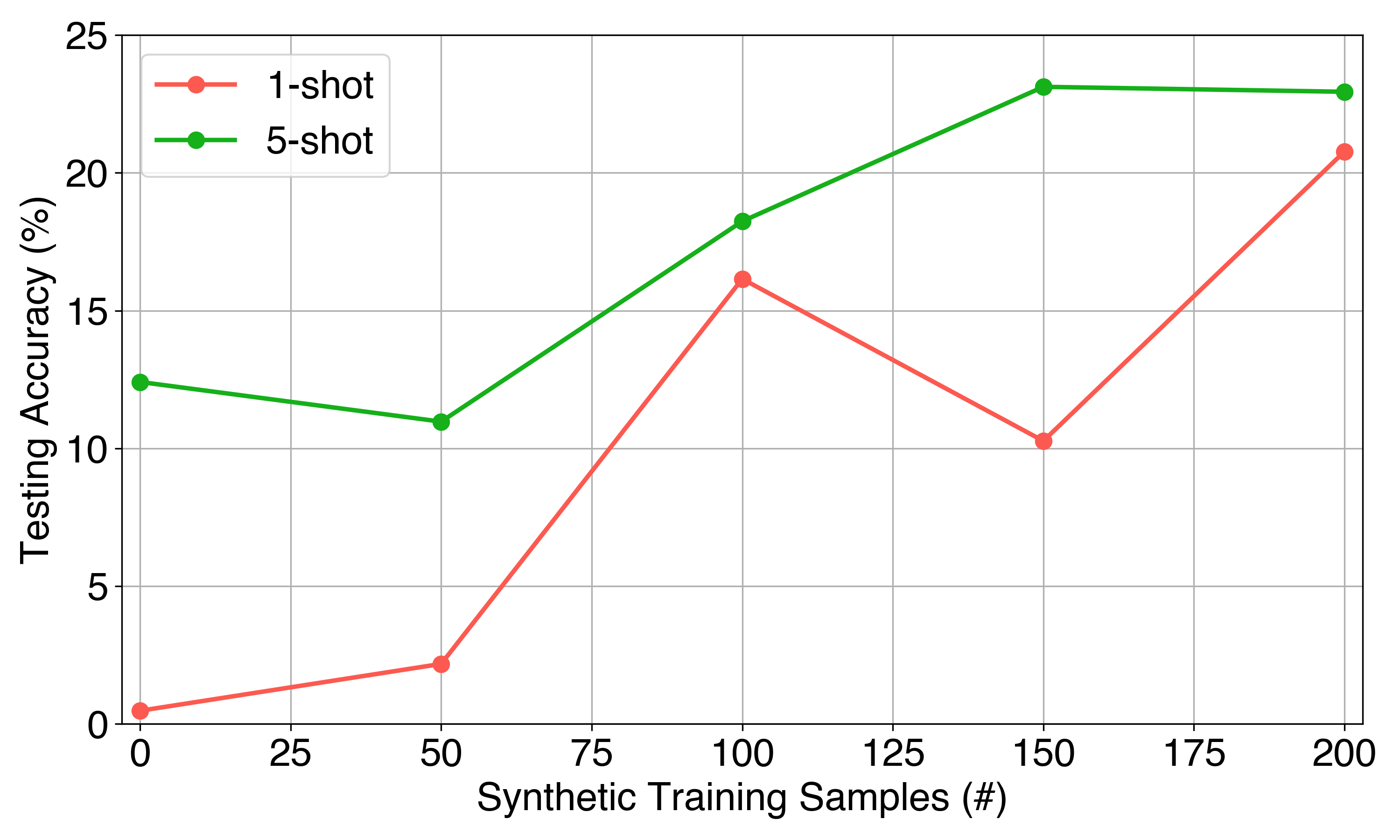}
    \caption{Testing accuracy on real videos from the Toyota Smarthome of two ResNet models trained in a one-shot (green) and few-shot (red) manner with an increasing amount of synthetic samples per class $n_{\text{background}}=0\dots200$ by steps of $50$.} 
    \label{fig:shots-toyota}
\end{figure}

\subsection{One-shot Experiments}

\begin{figure}
    \centering
    \includegraphics[width=\linewidth]{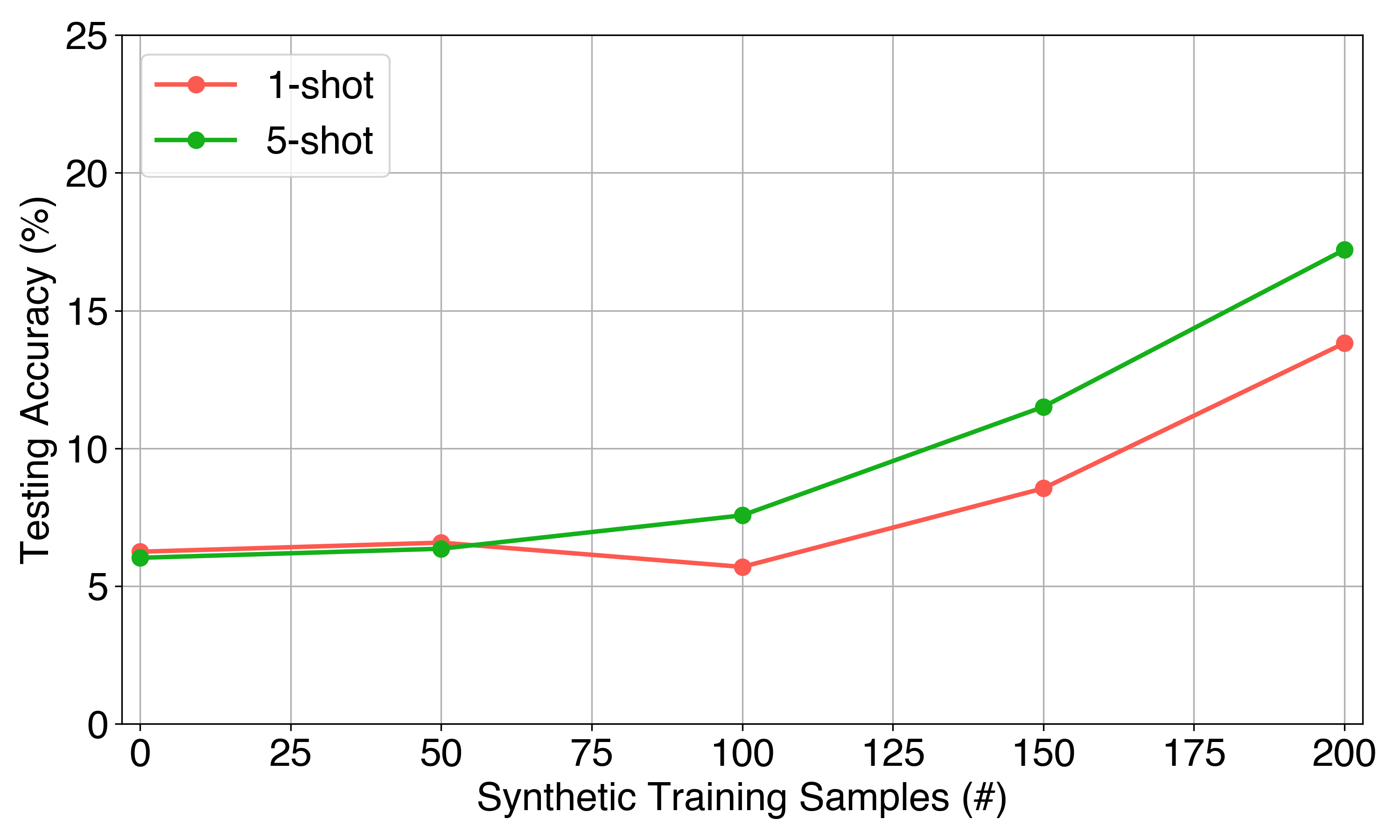}
    \caption{Testing accuracy on real videos from the NTU RGB+D of two ResNet models trained in a one-shot (green) and few-shot (red) manner with an increasing amount of synthetic samples per class $n_{\text{background}}=0\dots200$ by steps of $50$.}
    \label{fig:shots-ntu}
\end{figure}

Shown in Figures~\ref{fig:shots-toyota} and~\ref{fig:shots-ntu} (red curves) are the testing accuracy advantages gained when training one-shot models with our synthetic data on Toyota Smarthome and NTU RGB+D, respectively, plotted as a function of the number of synthetic samples.

With Toyota Smarthome, shown as the red curve in Figure~\ref{fig:shots-toyota}, the model starts below the accuracy of chance ($6.25\%$ for this $16$-class problem)\footnote{This was caused by a tendency for over-predicting one class. We repeated this experiment on different seeds but this phenomenon persisted.}. Once more than $100$ synthetic samples were included, the performance grew to $16\%$, $11\%$, and $21\%$, at $n_{\text{background}}=\{100,150,200\}$.

With NTU RGB+D, shown as the red curve in Figure~\ref{fig:shots-ntu}, the performance improved from $7\%$, the chance of luck, to $14\%$, doubling the original accuracy.

These results suggest that our synthetic data can meaningfully improve the performance of one-shot action recognition models. The trend observed in Figures~\ref{fig:shots-toyota} and~\ref{fig:shots-ntu} suggests that extending the synthetic data sample beyond $n_{\text{background}}=200$ may hold further improvement.

\subsection{Few-shot Experiments}

Shown in Figures~\ref{fig:shots-toyota} and~\ref{fig:shots-ntu} (green curves) are the testing accuracy advantages gained when training few-shot models with our synthetic data on Toyota Smarthome and NTU RGB+D, respectively, plotted as a function of the number of synthetic samples.

With Toyota Smarthome, shown as the green curve in Figure~\ref{fig:shots-toyota}, the performance improved from $13\%$ at $n_{\text{background}}=0$ to $23\%$ at $n_{\text{background}}=200$. 

With NTU RGB+D, shown as the green curve in Figure~\ref{fig:shots-ntu}, the performance improved from $7\%$, the chance of luck, at $n_{\text{background}}=0$ to $23\%$ at $n_{\text{background}}=200$.

These results indicate that our synthetic data can meaningfully improve the performance of few-shot action recognition models. The trend observed in Figure~\ref{fig:shots-toyota} suggests that, for Toyota Smarthome, the
performance gains saturate around $n_{\text{background}}=150$. For NTU RGB+D, shown in Figure~\ref{fig:shots-ntu}, there may be additional gains to be attained beyond $n_{\text{background}}=200$.

\section{Limitations}
\label{sec:limitations}

In this section, we enumerate the primary limitations of our method for synthetic video data generation. These limitations can be divided into two categories: those stemming from the employed pose transfer framework (in our case, ExAvatar) and those introduced directly by our method.

\subsection{Introduced by the Pose Transfer Framework}

The following limitations can be addressed in future work by utilizing newer pose transfer frameworks, which, we expect, will focus on addressing these shortcomings:

\textbf{L1.}~Videos with multiple people cannot be generated. This is because ExAvatar only supports pose transfer of videos with a single protagonist.

\textbf{L2.}~Actions that involve interaction with objects cannot be generated convincingly, as the interaction with the object will not be transferred. This is because ExAvatar does not support object interactions.



\subsection{Introduced by Our Method}

\begin{figure*}[t]
    \centering
    \includegraphics[width=0.16\textwidth]{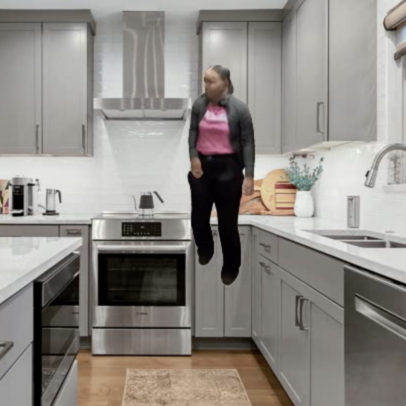}
    \hspace{0.01\textwidth}
    \includegraphics[width=0.16\textwidth]{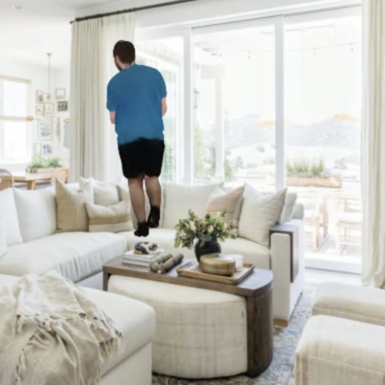}
    \hspace{0.01\textwidth}
    \includegraphics[width=0.16\textwidth]{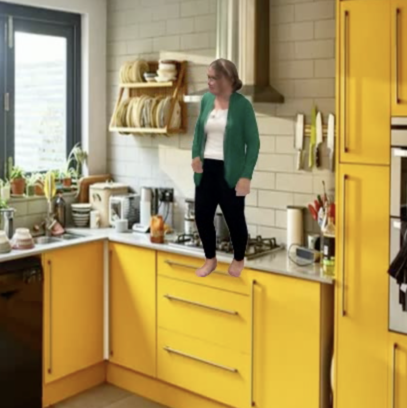}
    \hspace{0.01\textwidth}
    \includegraphics[width=0.16\textwidth]{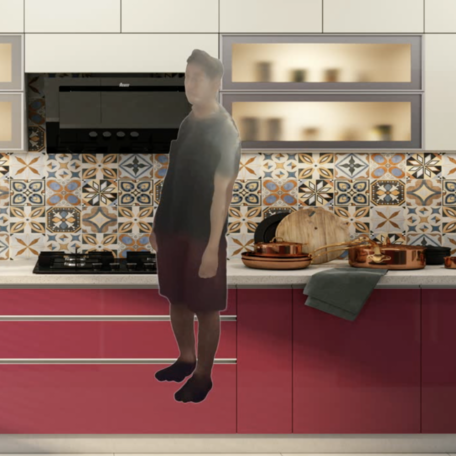}
    \hspace{0.01\textwidth}
    \includegraphics[width=0.16\textwidth]{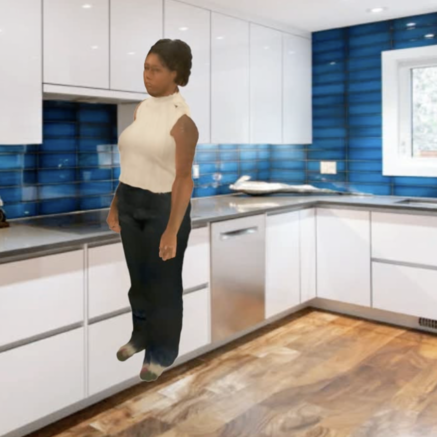}
    
    \caption{Example video frames illustrating limitation L3 (see Section~\ref{sec:limitations}).}
    \label{fig:L5_limitations_examples}
\end{figure*}

The following limitations can be addressed in future work by further improving our method (possible directions are discussed in Section~\ref{sec:discussion}):

\textbf{L3.}~In some videos generated with our method, the pose-transferred human action, superimposed atop the background image, may not be physically plausible in space. This is because our method currently does not reflect the semantics and depth element of the scene depicted in the background image. See examples in Figure~\ref{fig:L5_limitations_examples}.

\begin{figure*}[t]
    \centering
    \parbox{0.16\textwidth}{\centering\includegraphics[width=\linewidth]{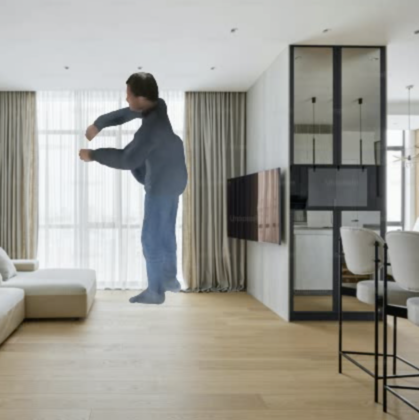}\\ \small Cook.stir}
    \hspace{0.01\textwidth}
    \parbox{0.16\textwidth}{\centering \includegraphics[width=\linewidth]{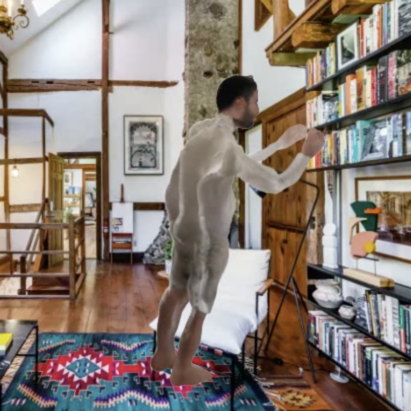}\\ \small Cook.Usestove}
    \hspace{0.01\textwidth}
    \parbox{0.16\textwidth}{\centering\includegraphics[width=\linewidth]{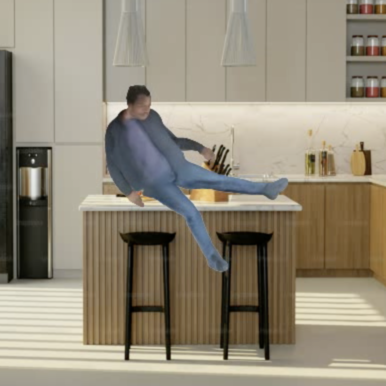}\\ \small Laydown}
    \hspace{0.01\textwidth}
    \parbox{0.16\textwidth}{\centering\includegraphics[width=\linewidth]{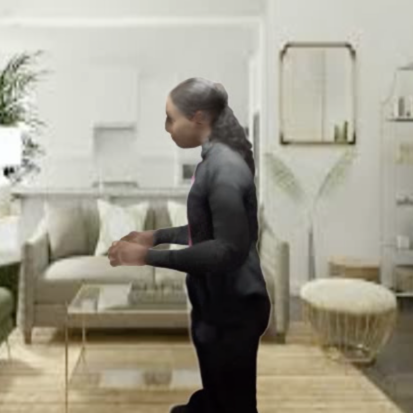}\\ \small Cook.cut}
    \hspace{0.01\textwidth}
    \parbox{0.16\textwidth}{\centering\includegraphics[width=\linewidth]{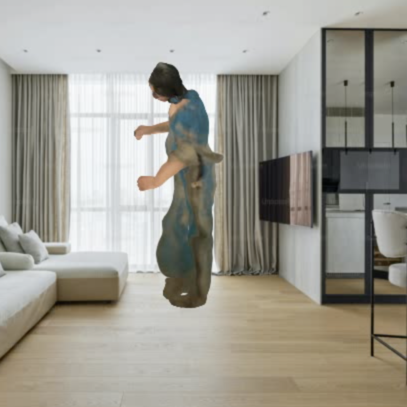}\\ \small Maketea.Insertteabag}
    
    \caption{Example video frames illustrating limitation L4 (see Section~\ref{sec:limitations}). The shown action classes are from Toyota Smarthome.}
    \label{fig:L6_limitation_examples}
\end{figure*}

\textbf{L4.}~Action classes that are associated with a particular setting may be generated in unnatural scenes or parts of scenes. Consider, for example, any cooking-related action classes (\textit{Cook: cut} or \textit{Cook: stir} in Toyota Smarthome), which would usually occur in the kitchen. Our method may generate new videos of these actions in inappropriate settings (e.g., the living room). See examples in Figure~\ref{fig:L6_limitation_examples}.

\section{Discussion}
\label{sec:discussion}

Our results demonstrate that the method presented in this paper enhances human action video classification model performance in both baseline, one-shot, and few-shot learning scenarios. By providing additional training samples that increase the identity diversity of the training data, our approach improves the model's ability to generalize from limited real data. Furthermore, it enhances background diversity, which proved particularly effective for NTU RGB+D, where all videos are captured in just two locations.

We expect this to be especially beneficial in applications where collecting large amounts of labeled video data is impractical or cost-prohibitive. Examples include sign language translation for low-resource sign languages and video processing for autonomous driving in challenging areas—such as cities and countries that have not been extensively mapped or scanned.

We believe that our method will play an important role in mitigating the bias of computer vision datasets and algorithms across various video understanding tasks. It has been shown that bias towards certain demographic groups in computer vision systems often lies in underrepresentation in the underlying training datasets~\cite{dehdashtian2024fairness,fabbrizzi2022survey}. This is where our method steps in: by generating new videos with protagonists of underrepresented demographic categories, it can balance such datasets.

Bias in computer vision systems is not limited to demographic groups, however. In the context of action recognition, for example, methods may suffer from a background bias, where the background is learned as a more predictive signal, leading the systems to ignore the actual human action~\cite{choi2019can}. Our method can be used to generate new training samples with varying backgrounds to mitigate this bias.

In terms of the employed pose transfer framework, we posit that methods combining 3D representations with recent generative AI techniques, similar to ExAvatar, will continue to be critical for synthetic video data generation. These approaches can leverage the fine-grained control and physical realism that come with 3D representations while enjoying the improved photorealism and scale of generative AI.

This, we believe, will also play an important role in addressing limitation L3: ensuring physical realism of the generated action in the new scene. Recent advancements in scene depth estimation~\cite{sayed2025doubletake,qin2024adaptive,lu2024self} will allow one to place the generated human motion (with a precise 3D representation) in a plausible part of the scene.

Limitation L4 may be addressed with image understanding techniques such as multimodal large language models, VQA models, or multimodal embeddings. These could quantify the appropriateness of a scene for a particular action.
\section{Conclusion}
\label{sec:conclusion}

In this paper, we introduce a novel framework for synthetic data generation of human action video. By leveraging a modified ExAvatar framework for 3D Gaussian avatar animation, our method reenacts human actions from reference videos using novel human identities in varied settings. By combining computer graphics and generative AI frameworks, this approach addresses limitations in photorealism and semantic control that have hindered previous synthetic data generation methods for video understanding tasks.

We evaluate our method on a subset of the Toyota Smarthome and NTU RGB+D dataset, demonstrating notable improvements in action recognition performance across two video classification architectures. In particular, our method yields significant improvements in baseline, one-shot, and few-shot learning scenarios.

Finally, we present the \textit{RANDOM People} dataset, which contains synthetic videos, novel human identity videos along with their avatars, and scene background images. 

The dataset, code, and additional resources are open-sourced at \url{synthetic-human-action.github.io}. In future work, we aim to resolve limitations L3 and L4 pertaining suitable background selection and adjustment, extend our system's ability to capture and render human interactions with objects and further improve the photorealism of the resulting videos.

{
    \small
    \bibliographystyle{ieeenat_fullname}
    \bibliography{main}
}

\clearpage
\setcounter{page}{1}
\maketitlesupplementary

\setcounter{section}{0}

\section{Prolific Participant Instructions}
\label{app:participant_instructions}

As described in Section~\ref{subsec:data__novel_human_identity_videos}, to create the \textit{RANDOM People} dataset, we crowd-sourced novel human identity videos using the Prolific data platform. Before entering the recording interface and seeing any instructions, the participants were informed about the intended use of the dataset, and asked whether they consent to their video---as well as a 3D Gaussian model and other derivate artifacts---being publicly available for research purposes.

Once the users agreed, they advanced to the recording interface, where they were presented with a video showing the action to perform and the following instructions:

\begin{tcolorbox}[colback=lightgray, colframe=black, title=Participant Instructions]

Watch [this video] on YouTube. You will use your phone or tablet to record yourself performing the same sequence of actions. \\

\textbf{First, prepare the recording.} Place your phone or tablet approximately 7-8 feet (2-2.5 meters) away on an elevated surface. The phone should be positioned at a height above your waist level. Ensure that you are fully visible and approximately in the center of the frame.

\textbf{Next, proceed with recording} yourself while performing the following sequence of actions:

\begin{enumerate}
    \item a slow 360 rotation with your hands down;
    \item a slow 360 rotation with your hands up in a double L shape as shown below;
    \item a slow 360 rotation with your hands down.
\end{enumerate}

\textbf{Importantly}, the recording must meet the following \textbf{criteria}:

\begin{itemize}
\item Your whole body, head to feet, is visible in the video at all times.
\item You must be well-lit.
\item Besides you, no other people, animals, or moving objects appear in the video. This includes statues, posters, and TV.
\item Your camera is positioned on an elevated surface, such as a table or wardrobe—do not record with a phone placed on the ground.  \\
\end{itemize} 

When you're ready, upload the video below. By uploading, you agree to [these terms].

Thank you!

\end{tcolorbox}

\section{Selected Action Classes}
\label{app:action_classes}

As described in Section~\ref{sec:data}, we manually selected a subset of $16$ action classes within the Toyota Smarthome~\cite{Dai_2022_PAMI} and NTU RGB-D~\cite{shahroudy2016ntu} based on the following criteria: (1) Minimal Use of External Objects, (2) Consistent Camera Angles, and (3) Distinctive Actions. In particular, these subsets include:

\begin{tcolorbox}[colback=lightgreen, colframe=black, title=Selected Action Classes: Toyota Smarthome]
    \begin{enumerate}
        \item Cook.cut
        \item Cook.stir
        \item Cook.Usestove
        \item Drink.Frombottle
        \item Drink.Fromcan
        \item Drink.Fromcup
        \item Eat.snack
        \item Getup
        \item Laydown
        \item Pour.Fromkettle
        \item Pour.Frombottle
        \item Sitdown
        \item Walk
        \item Usetelephone
        \item Maketea.Insertteabag
        \item Enter
    \end{enumerate}
\end{tcolorbox}

\begin{tcolorbox}[colback=lightgreen, colframe=black, title=Selected Action Classes: NTU RGB-D]
    \begin{enumerate}
        \item drink water (A1)
        \item eat meal (A2)
        \item brush teeth (A3)
        \item pick up (A6)
        \item throw (A7)
        \item sit down (A8)
        \item stand up (A9)
        \item clapping (A10)
        \item hand waving (A23)
        \item kicking something (A24)
        \item jump up (A27)
        \item point to something (A31)
        \item nod head/bow (A35)
        \item salute (A38)
        \item put palms together (A39)
        \item cross hands in front (A40)
    \end{enumerate}
\end{tcolorbox}

\section{Compute Considerations}
\label{app:infra_and_runtime_considerations}

This appendix section discusses compute considerations surrounding our experimental setup. Our aim is to provide an intuition for the computational demands of this process and to explain the parameters we chose, which were largely constrained by our computing capacity. Due to limited GPU access, we were only able to perform the experiments on the \textit{RANDOM People 15} subset with 15 novel human identities instead of the complete set of 100 novel human identities.

These identity videos $I$ were standardized to $18$ seconds at $18$ FPS; the reference videos $T$ were normalized to $20$ seconds at $25$ FPS. While the statistics reported below, which informed this parameter choice, have been measured precisely, this is not meant to constitute a formal analysis of the running time and optimization; rather, we aim to equip the reader with an understanding of the approximate computing complexity and the rationale behind our parameter decisions.

Most identity videos in $I$ collected for \textit{RANDOM People} were between $40$ to $60$ seconds in length, containing approximately $1,200$ frames. Creating an avatar (as described in Section~\ref{sec:methods}) from a single identity video in $I$ on an NVIDIA RTX 4090 GPU took approximately six hours. By normalizing the videos to $18$ seconds at $18$ FPS, the avatar creation time was reduced by a factor of four, down to approximately $1.5$ hours.

We explored additional configurations as well. When normalizing to $20$ seconds at $25$ FPS, the processing time was approximately $2.5$ hours. At $20$ seconds and $20$ FPS, the processing time was around $2.3$ hours, and at $20$ seconds and $18$ FPS, it was roughly $1.8$ hours.

However, when reducing the frame count further, we observed a decline in the quality of the final avatar. Ultimately, we found that the optimal balance between model accuracy and processing time was achieved with approximately $320$ training frames per identity.

\newpage

\section{Qualitative Evaluation}
\label{app:qualitative_eval}

\begin{figure*}[h]
    \centering
    \renewcommand{\arraystretch}{2.5}

    \begin{tabular}{@{}c@{\hskip 10pt}c@{\hskip 10pt}c@{\hskip 10pt}c@{\hskip 10pt}c@{}}
        \makebox[0.16\textwidth]{$t=0$} & 
        \makebox[0.16\textwidth]{$t=1$} & 
        \makebox[0.16\textwidth]{$t=2$} & 
        \makebox[0.16\textwidth]{$t=3$} & 
        \makebox[0.16\textwidth]{$t=4$} \\
    \end{tabular}

    \vspace{2pt}

    \fbox{
        \begin{tabular}{@{}c@{\hskip 10pt}c@{\hskip 10pt}c@{\hskip 10pt}c@{\hskip 10pt}c@{}}
            \includegraphics[width=0.16\textwidth]{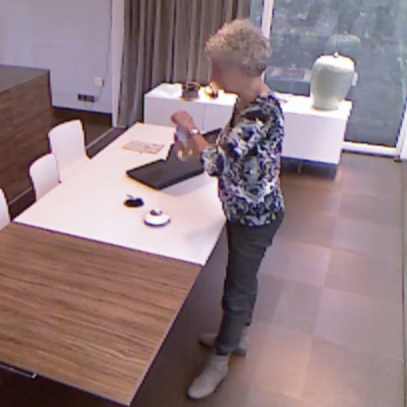} &
            \includegraphics[width=0.16\textwidth]{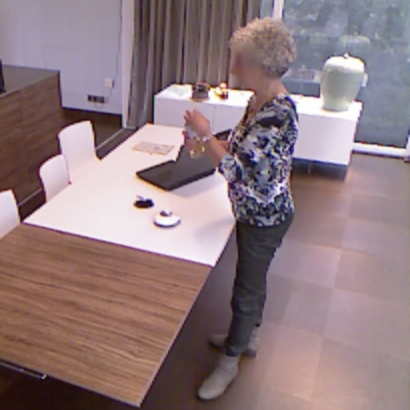} &
            \includegraphics[width=0.16\textwidth]{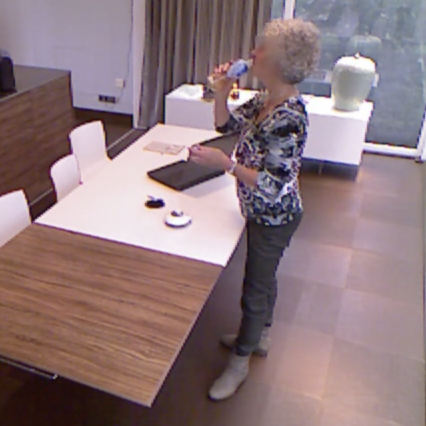} &
            \includegraphics[width=0.16\textwidth]{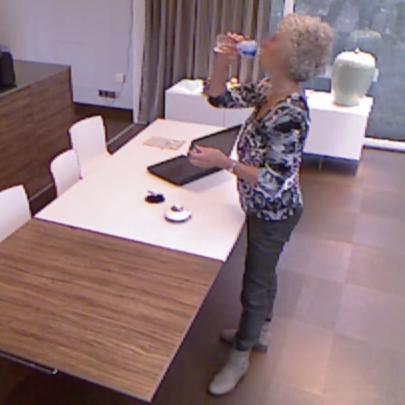} &
            \includegraphics[width=0.16\textwidth]{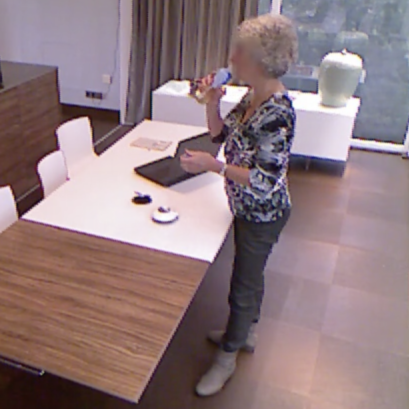} \\
    
            \includegraphics[width=0.16\textwidth]{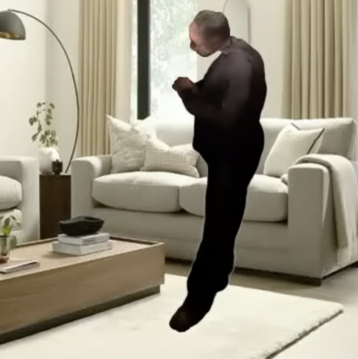} &
            \includegraphics[width=0.16\textwidth]{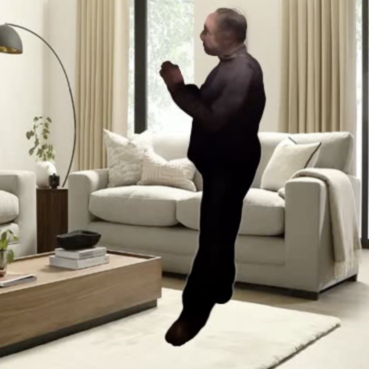} &
            \includegraphics[width=0.16\textwidth]{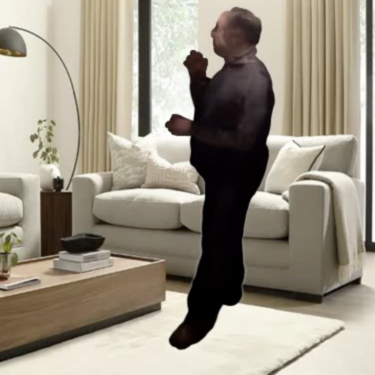} &
            \includegraphics[width=0.16\textwidth]{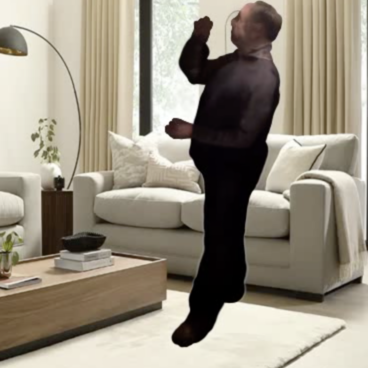} &
            \includegraphics[width=0.16\textwidth]{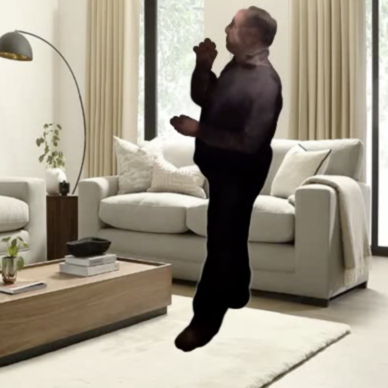} \\
        \end{tabular}
    }

    \vspace{0.3cm}

    \fbox{
        \begin{tabular}{@{}c@{\hskip 10pt}c@{\hskip 10pt}c@{\hskip 10pt}c@{\hskip 10pt}c@{}}
            \includegraphics[width=0.16\textwidth]{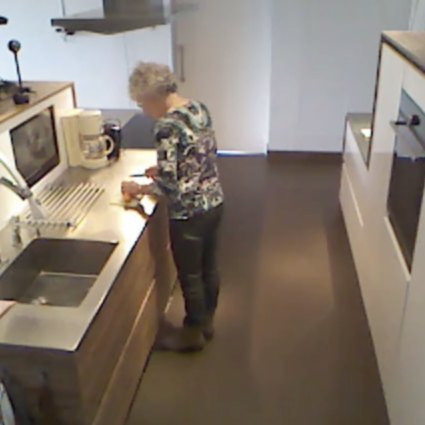} &
            \includegraphics[width=0.16\textwidth]{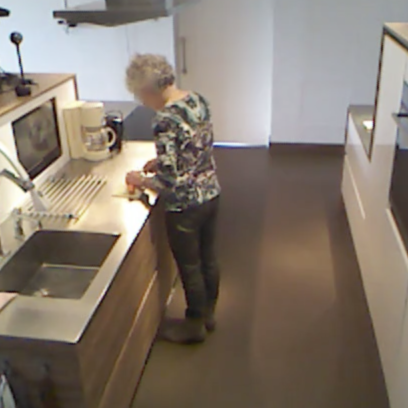} &
            \includegraphics[width=0.16\textwidth]{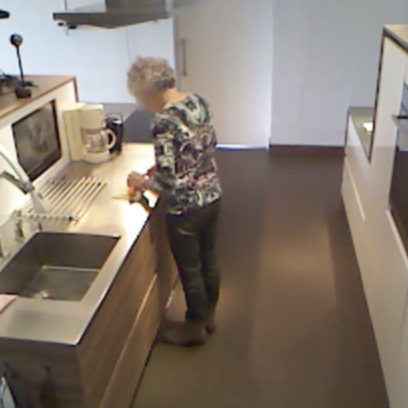} &
            \includegraphics[width=0.16\textwidth]{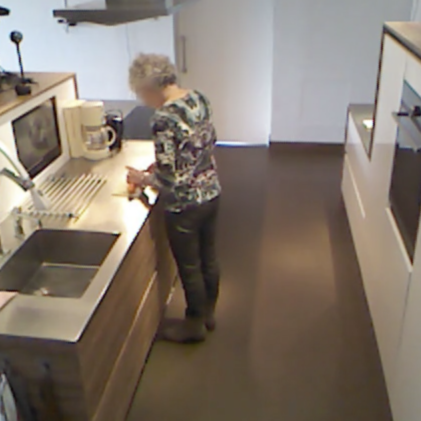} &
            \includegraphics[width=0.16\textwidth]{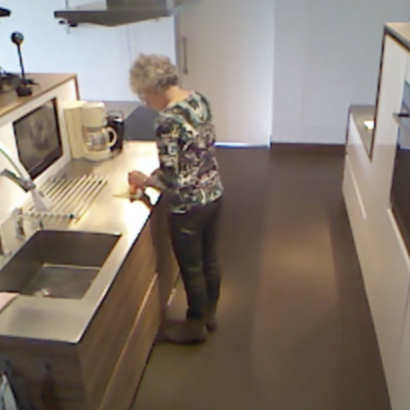} \\
    
            \includegraphics[width=0.16\textwidth]{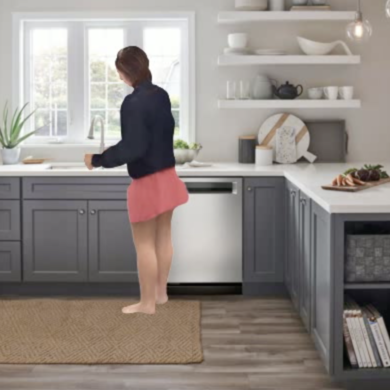} &
            \includegraphics[width=0.16\textwidth]{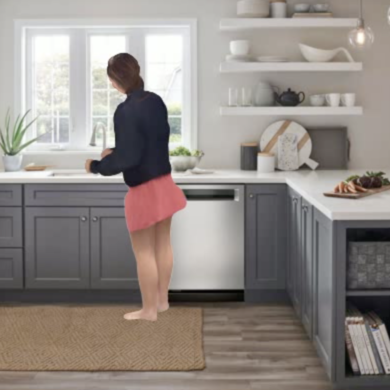} &
            \includegraphics[width=0.16\textwidth]{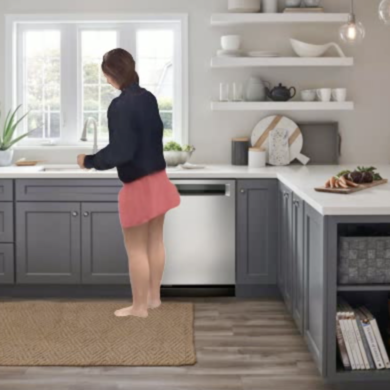} &
            \includegraphics[width=0.16\textwidth]{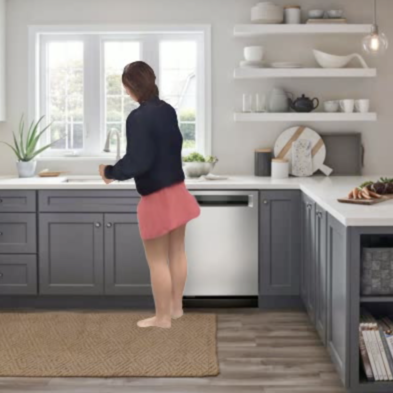} &
            \includegraphics[width=0.16\textwidth]{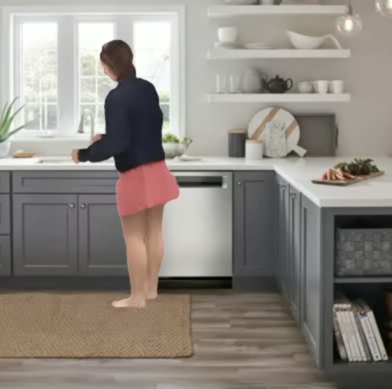} \\
        \end{tabular}
    }

    \vspace{0.3cm}

    \fbox{
        \begin{tabular}{@{}c@{\hskip 10pt}c@{\hskip 10pt}c@{\hskip 10pt}c@{\hskip 10pt}c@{}}
            \includegraphics[width=0.16\textwidth]{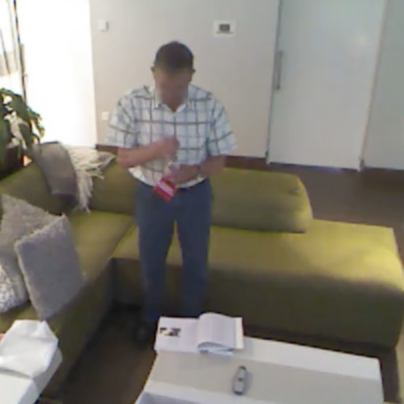} &
            \includegraphics[width=0.16\textwidth]{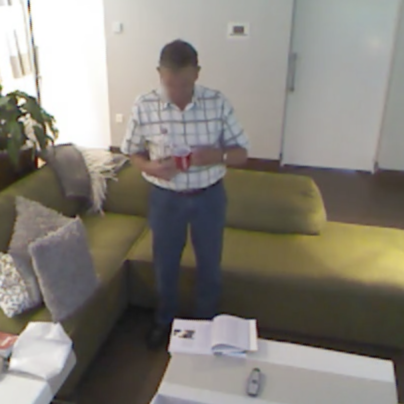} &
            \includegraphics[width=0.16\textwidth]{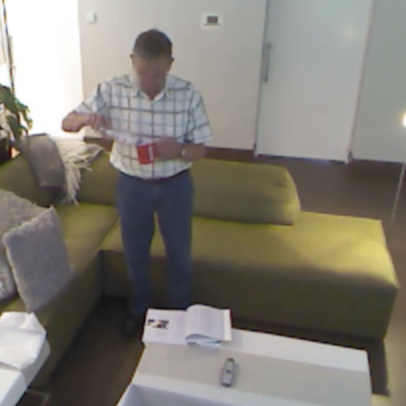} &
            \includegraphics[width=0.16\textwidth]{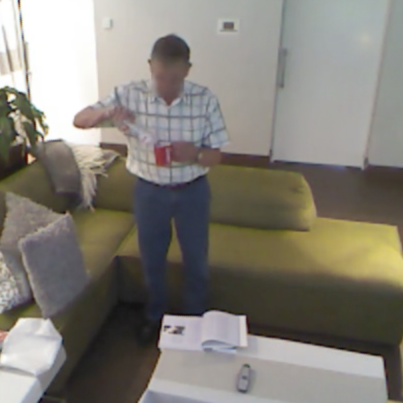} &
            \includegraphics[width=0.16\textwidth]{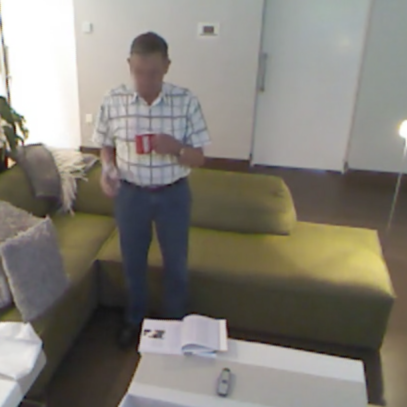} \\
    
            \includegraphics[width=0.16\textwidth]{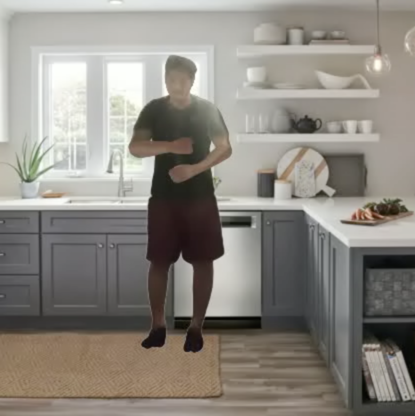} &
            \includegraphics[width=0.16\textwidth]{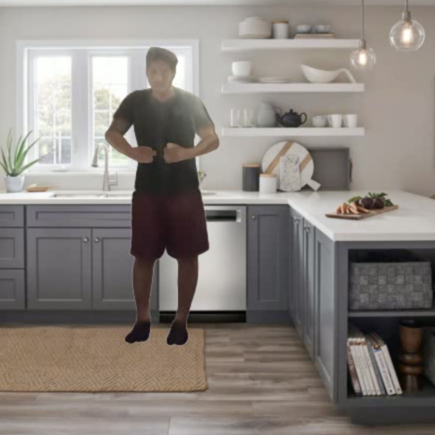} &
            \includegraphics[width=0.16\textwidth]{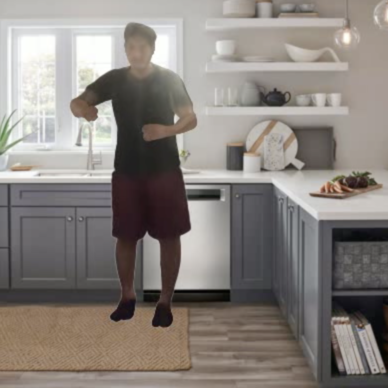} &
            \includegraphics[width=0.16\textwidth]{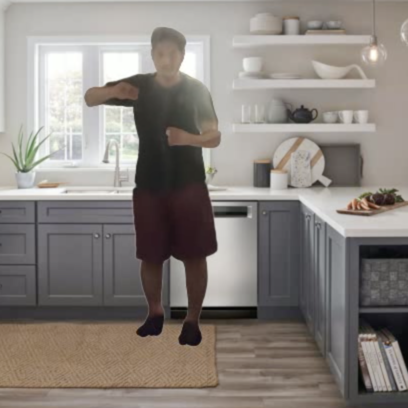} &
            \includegraphics[width=0.16\textwidth]{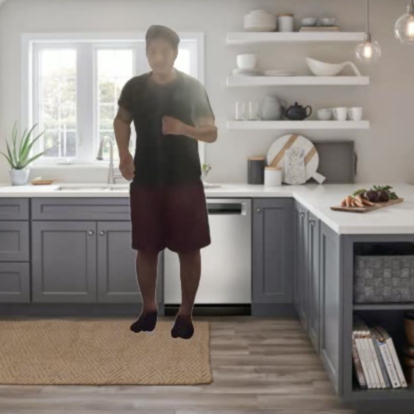} \\
        \end{tabular}
    }

    \caption{Examples of video frames at $t=\{0, 1, 2, 3, 4\}$ seconds from the source video (top), taken from Toyota Smarthome, and the target video (bottom), generated by our synthetic data generation method, where the pose alignment is consistent.}
    \label{fig:consistency_frame_grid_consistent}
\end{figure*}

\begin{figure*}[h]
    \centering
    \renewcommand{\arraystretch}{2.5}

    \begin{tabular}{@{}c@{\hskip 10pt}c@{\hskip 10pt}c@{\hskip 10pt}c@{\hskip 10pt}c@{}}
        \makebox[0.16\textwidth]{$t=0$} & 
        \makebox[0.16\textwidth]{$t=1$} & 
        \makebox[0.16\textwidth]{$t=2$} & 
        \makebox[0.16\textwidth]{$t=3$} & 
        \makebox[0.16\textwidth]{$t=4$} \\
    \end{tabular}

    \vspace{2pt}

    \fbox{
        \begin{tabular}{@{}c@{\hskip 10pt}c@{\hskip 10pt}c@{\hskip 10pt}c@{\hskip 10pt}c@{}}
        \includegraphics[width=0.16\textwidth]{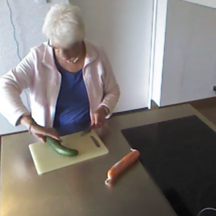} &
        \includegraphics[width=0.16\textwidth]{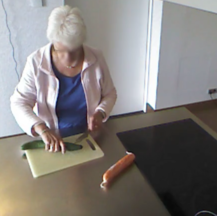} &
        \includegraphics[width=0.16\textwidth]{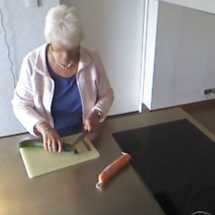} &
        \includegraphics[width=0.16\textwidth]{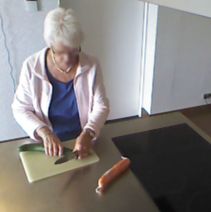} &
        \includegraphics[width=0.16\textwidth]{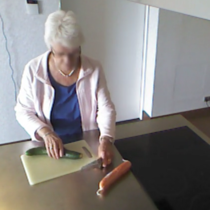} \\

        \includegraphics[width=0.16\textwidth]{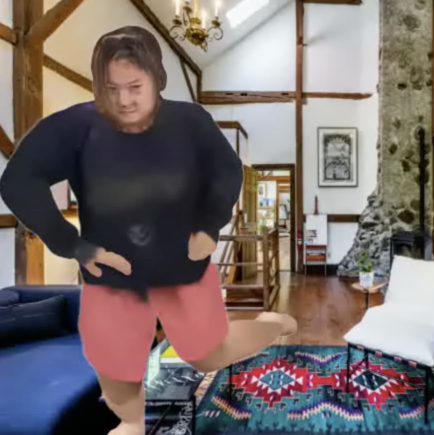} &
        \includegraphics[width=0.16\textwidth]{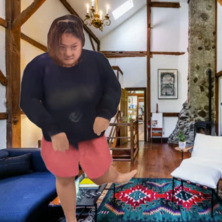} &
        \includegraphics[width=0.16\textwidth]{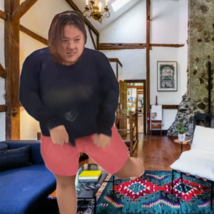} &
        \includegraphics[width=0.16\textwidth]{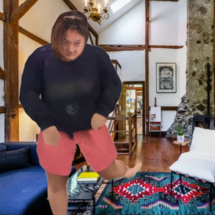} &
        \includegraphics[width=0.16\textwidth]{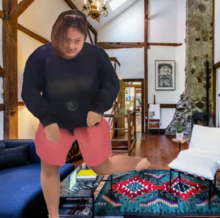} \\
        \end{tabular}
    }

    \vspace{0.3cm}

    \fbox{
        \begin{tabular}{@{}c@{\hskip 10pt}c@{\hskip 10pt}c@{\hskip 10pt}c@{\hskip 10pt}c@{}}
            \includegraphics[width=0.16\textwidth]{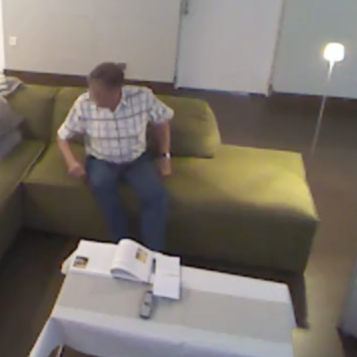} &
            \includegraphics[width=0.16\textwidth]{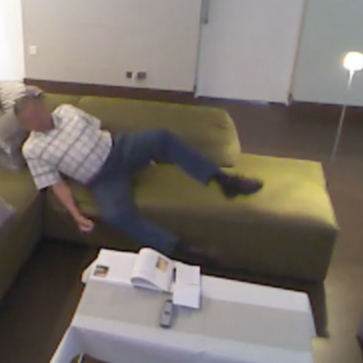} &
            \includegraphics[width=0.16\textwidth]{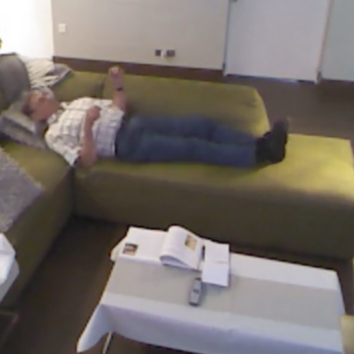} &
            \includegraphics[width=0.16\textwidth]{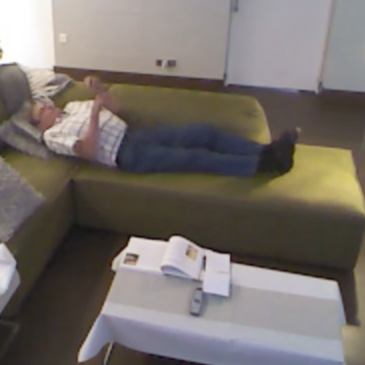} &
            \includegraphics[width=0.16\textwidth]{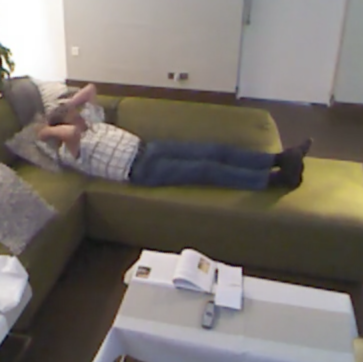} \\
    
            \includegraphics[width=0.16\textwidth]{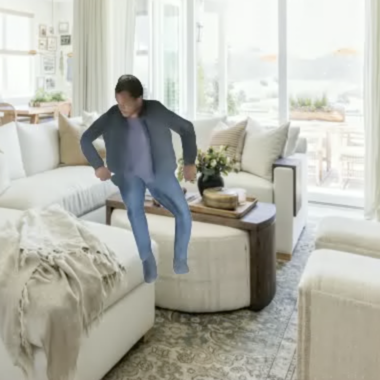} &
            \includegraphics[width=0.16\textwidth]{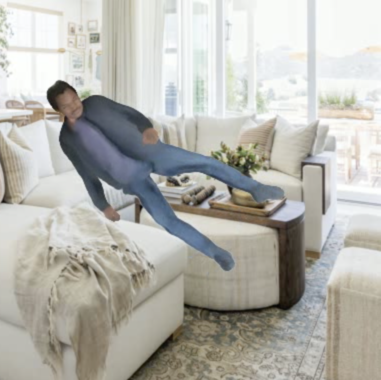} &
            \includegraphics[width=0.16\textwidth]{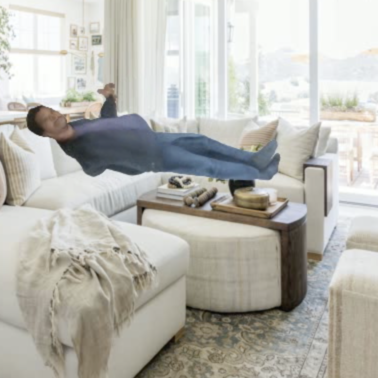} &
            \includegraphics[width=0.16\textwidth]{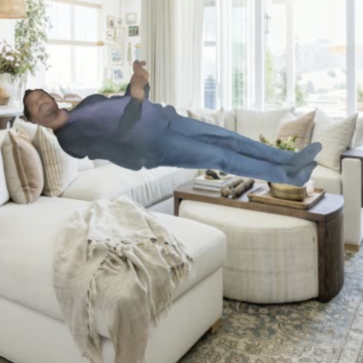} &
            \includegraphics[width=0.16\textwidth]{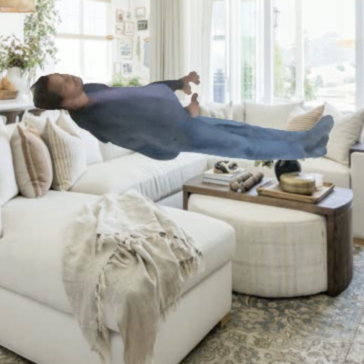} \\
        \end{tabular}
    }

    \vspace{0.3cm}

    \fbox{
        \begin{tabular}{@{}c@{\hskip 10pt}c@{\hskip 10pt}c@{\hskip 10pt}c@{\hskip 10pt}c@{}}
            \includegraphics[width=0.16\textwidth]{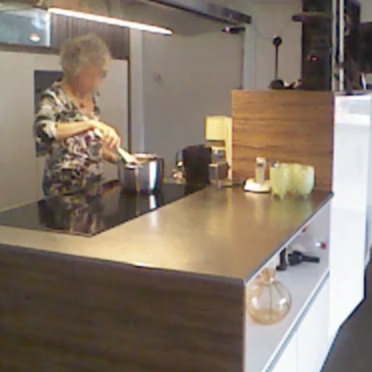} &
            \includegraphics[width=0.16\textwidth]{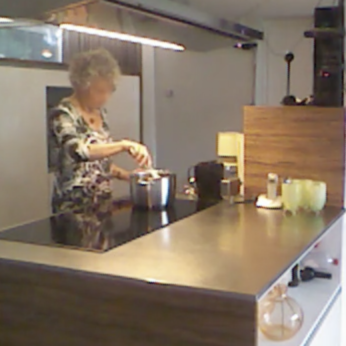} &
            \includegraphics[width=0.16\textwidth]{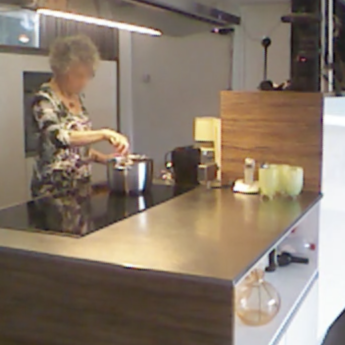} &
            \includegraphics[width=0.16\textwidth]{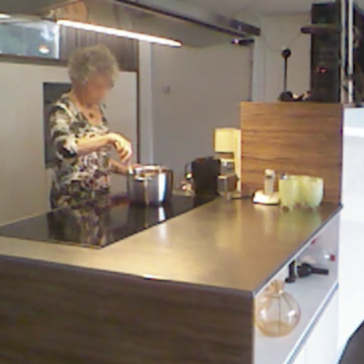} &
            \includegraphics[width=0.16\textwidth]{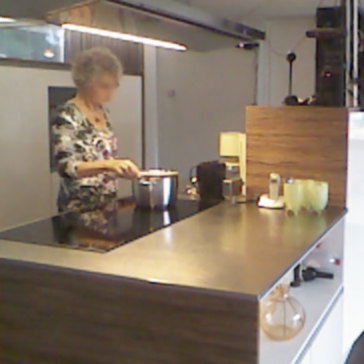} \\
    
            \includegraphics[width=0.16\textwidth]{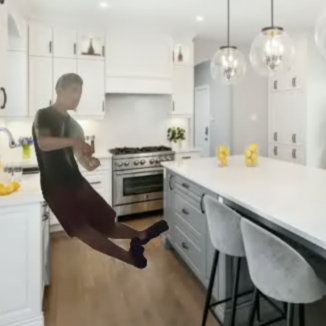} &
            \includegraphics[width=0.16\textwidth]{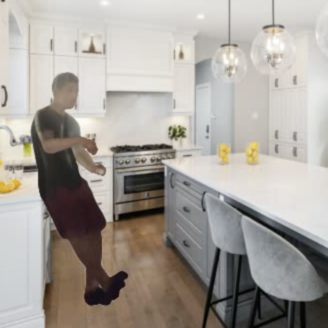} &
            \includegraphics[width=0.16\textwidth]{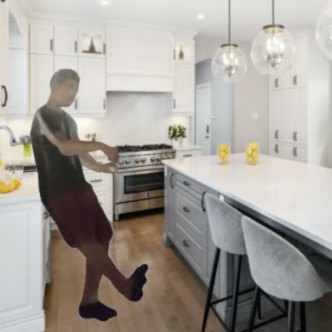} &
            \includegraphics[width=0.16\textwidth]{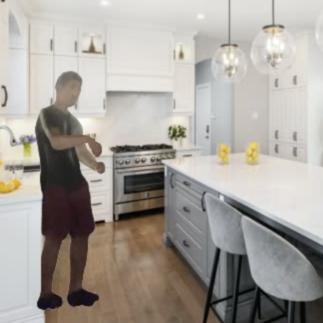} &
            \includegraphics[width=0.16\textwidth]{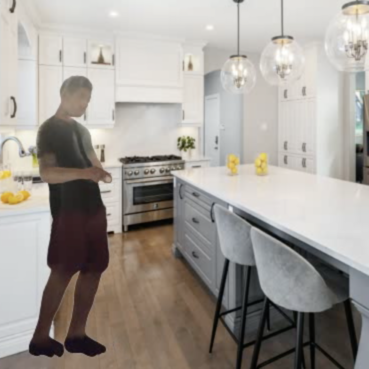} \\
        \end{tabular}
    }

    \caption{Examples of video frames at $t=\{0, 1, 2, 3, 4\}$ seconds from the source video (top), taken from Toyota Smarthome, and the target video (bottom), generated by our synthetic data generation method, where the pose alignment is inconsistent.}
    \label{fig:consistency_frame_grid_inconsistent}
\end{figure*}
\begin{figure*}[h]
    \centering
    \renewcommand{\arraystretch}{2.5}

    \begin{tabular}{@{}c@{\hskip 10pt}c@{\hskip 10pt}c@{\hskip 10pt}c@{\hskip 10pt}c@{}}
        \makebox[0.16\textwidth]{$t=0$} & 
        \makebox[0.16\textwidth]{$t=1$} & 
        \makebox[0.16\textwidth]{$t=2$} & 
        \makebox[0.16\textwidth]{$t=3$} & 
        \makebox[0.16\textwidth]{$t=4$} \\
    \end{tabular}

    \vspace{2pt}

    \fbox{
        \begin{tabular}{@{}c@{\hskip 10pt}c@{\hskip 10pt}c@{\hskip 10pt}c@{\hskip 10pt}c@{}}
            \includegraphics[width=0.16\textwidth]{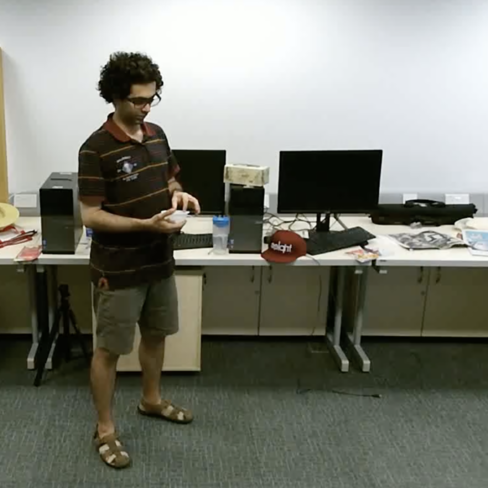} &
            \includegraphics[width=0.16\textwidth]{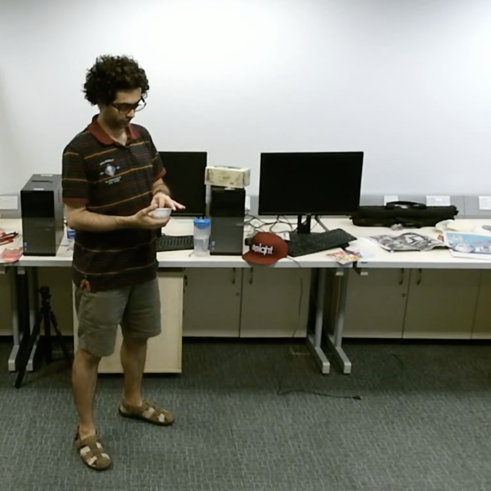} &
            \includegraphics[width=0.16\textwidth]{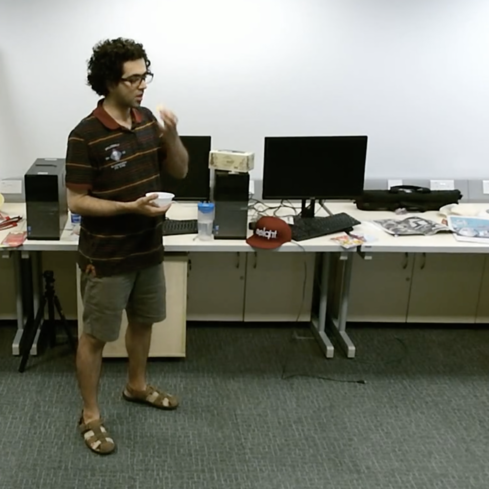} &
            \includegraphics[width=0.16\textwidth]{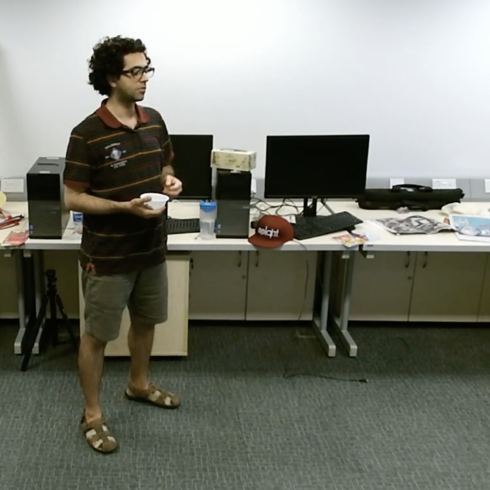} &
            \includegraphics[width=0.16\textwidth]{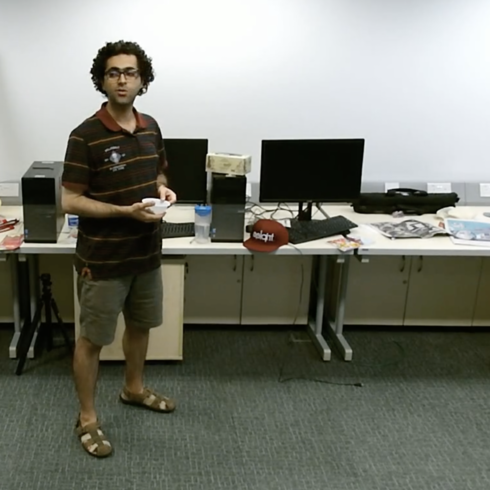} \\
    
            \includegraphics[width=0.16\textwidth]{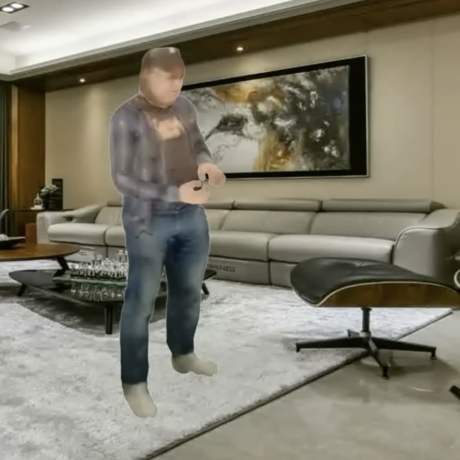} &
            \includegraphics[width=0.16\textwidth]{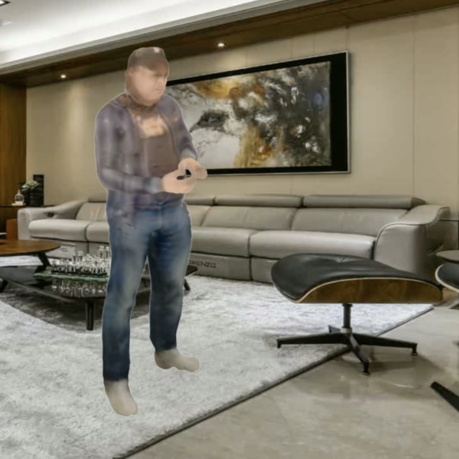} &
            \includegraphics[width=0.16\textwidth]{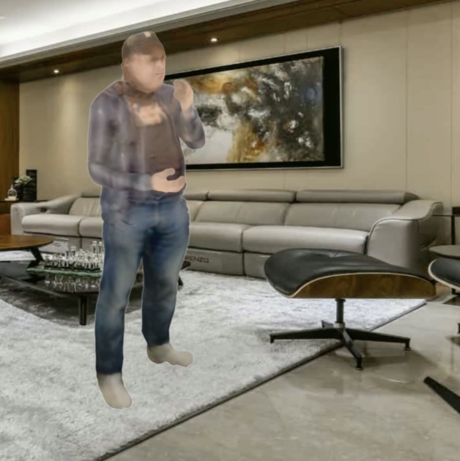} &
            \includegraphics[width=0.16\textwidth]{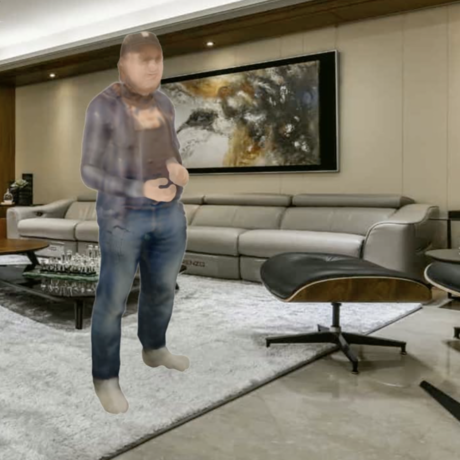} &
            \includegraphics[width=0.16\textwidth]{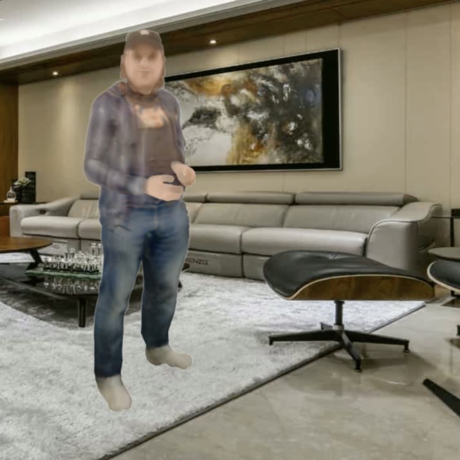} \\
        \end{tabular}
    }

    \vspace{0.3cm}

    \fbox{
        \begin{tabular}{@{}c@{\hskip 10pt}c@{\hskip 10pt}c@{\hskip 10pt}c@{\hskip 10pt}c@{}}
            \includegraphics[width=0.16\textwidth]{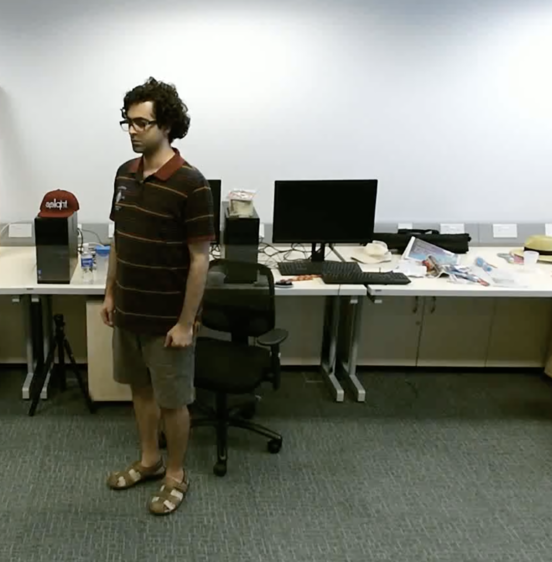} &
            \includegraphics[width=0.16\textwidth]{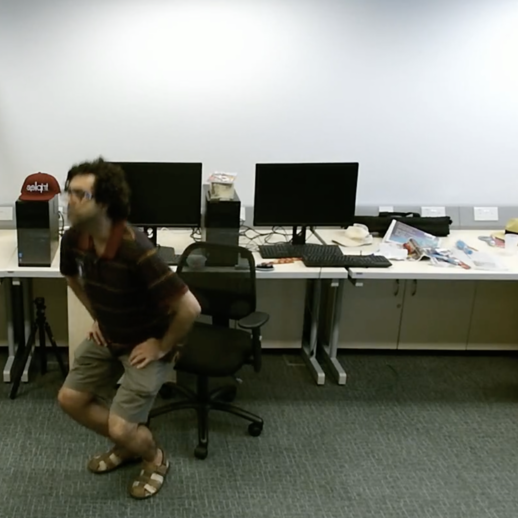} &
            \includegraphics[width=0.16\textwidth]{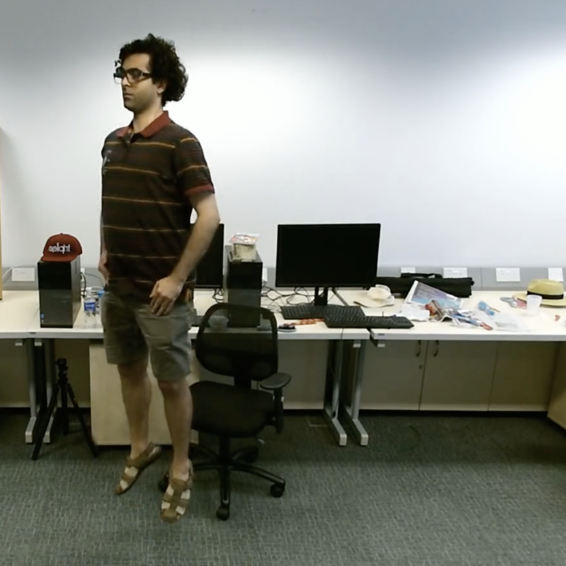} &
            \includegraphics[width=0.16\textwidth]{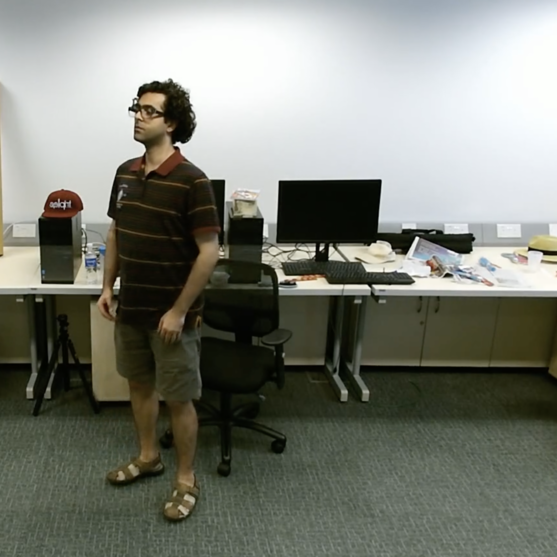} &
            \includegraphics[width=0.16\textwidth]{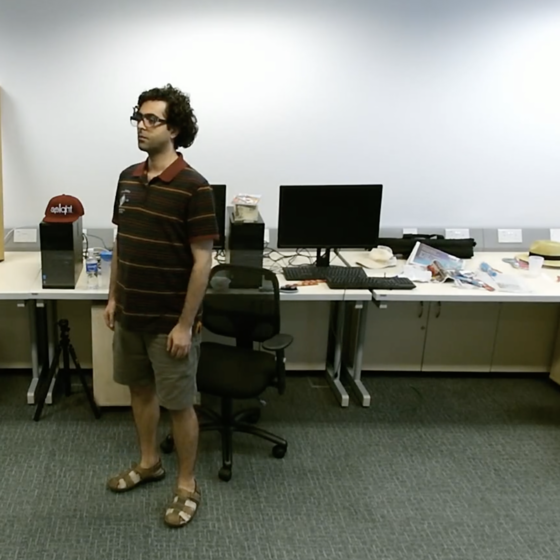} \\
    
            \includegraphics[width=0.16\textwidth]{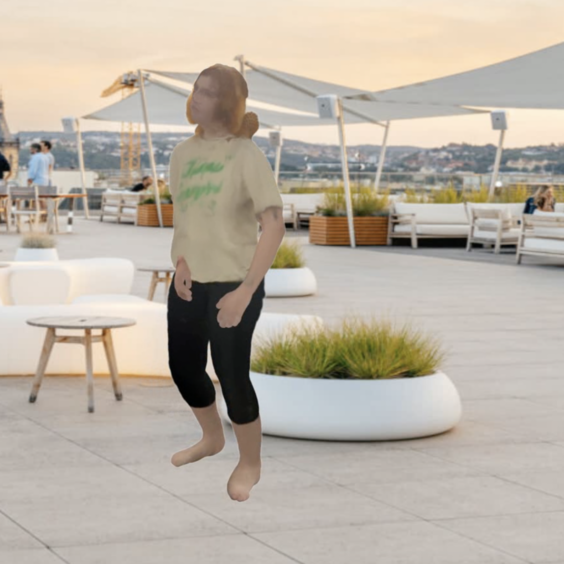} &
            \includegraphics[width=0.16\textwidth]{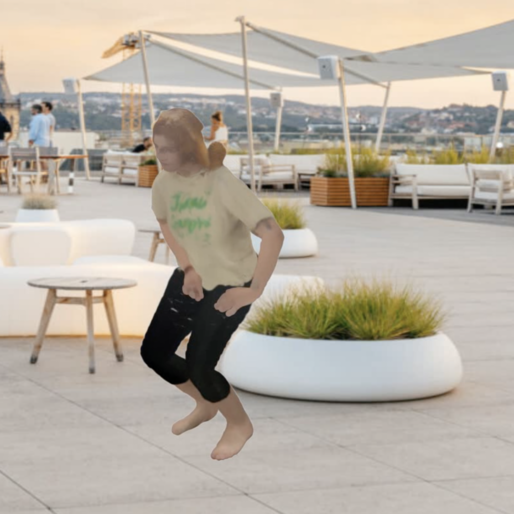} &
            \includegraphics[width=0.16\textwidth]{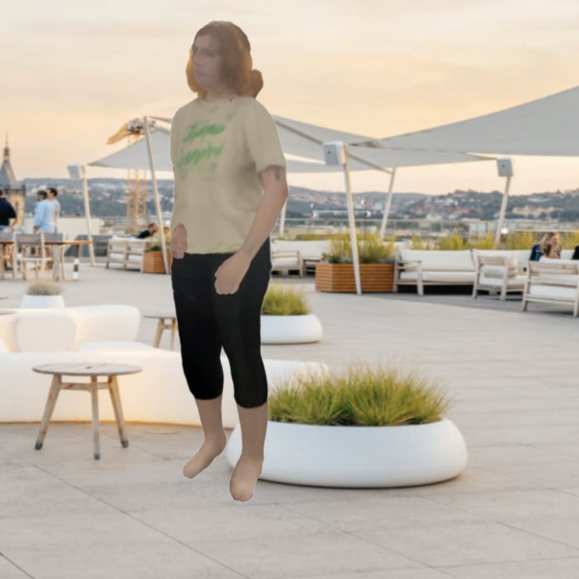} &
            \includegraphics[width=0.16\textwidth]{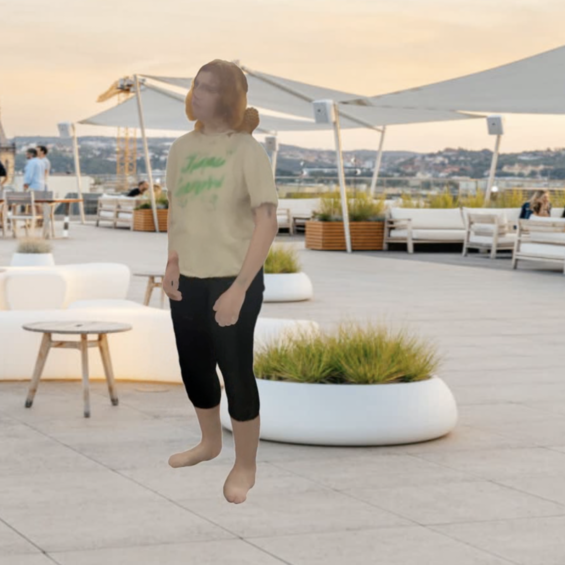} &
            \includegraphics[width=0.16\textwidth]{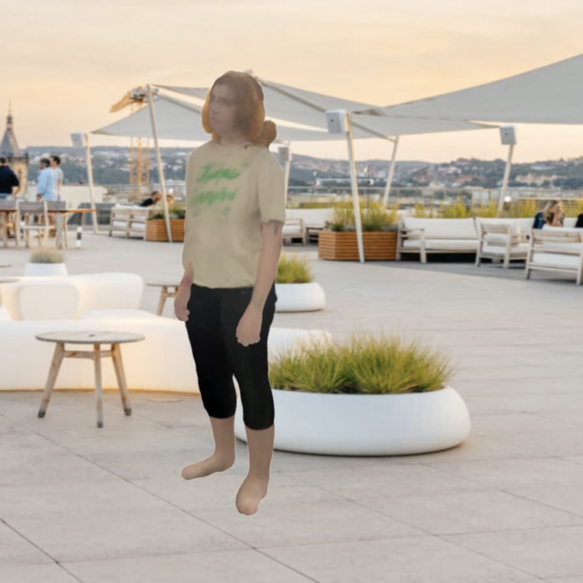} \\
        \end{tabular}
    }

    \vspace{0.3cm}

    \fbox{
        \begin{tabular}{@{}c@{\hskip 10pt}c@{\hskip 10pt}c@{\hskip 10pt}c@{\hskip 10pt}c@{}}
            \includegraphics[width=0.16\textwidth]{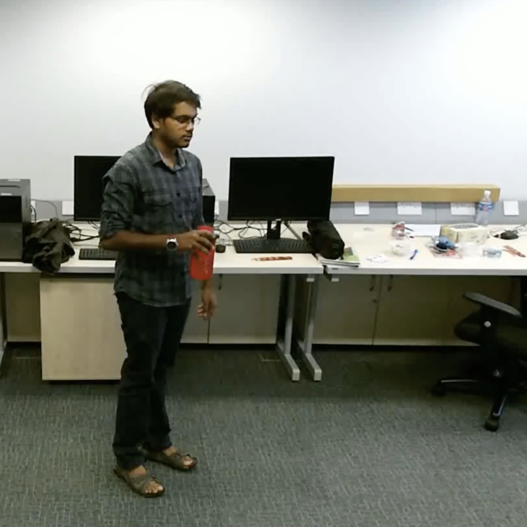} &
            \includegraphics[width=0.16\textwidth]{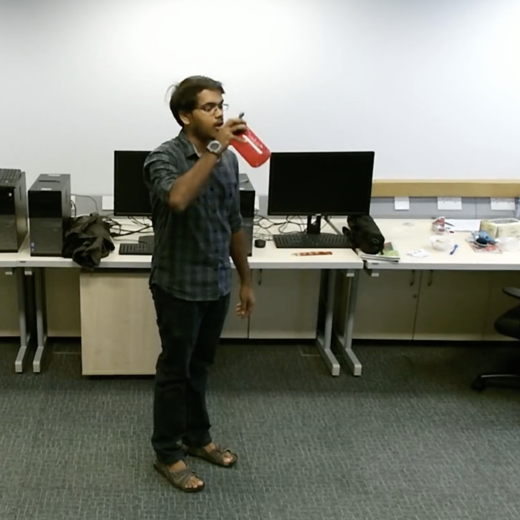} &
            \includegraphics[width=0.16\textwidth]{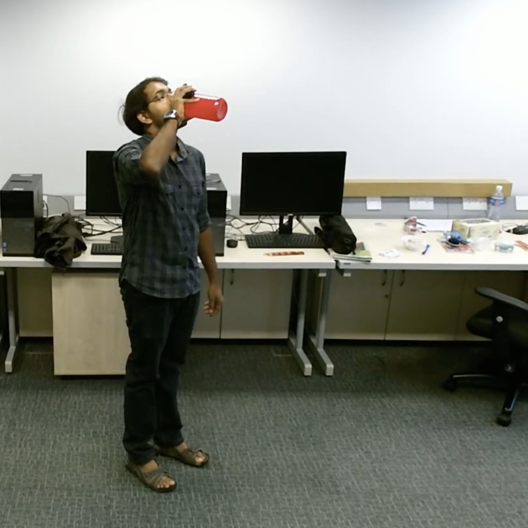} &
            \includegraphics[width=0.16\textwidth]{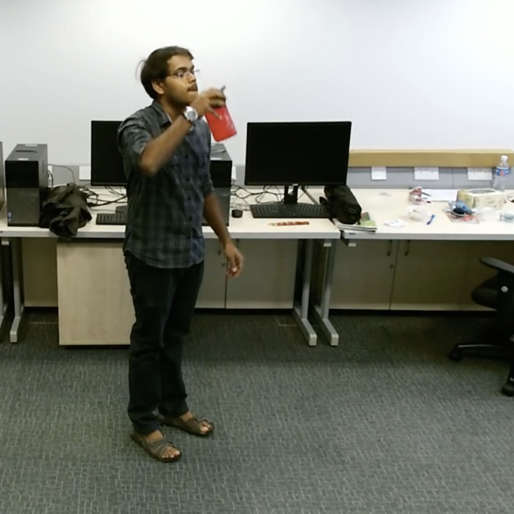} &
            \includegraphics[width=0.16\textwidth]{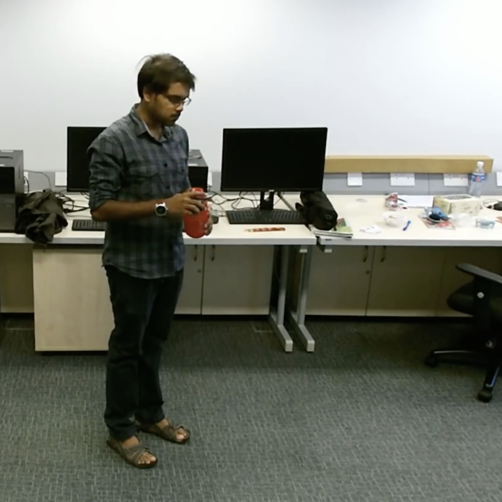} \\
    
            \includegraphics[width=0.16\textwidth]{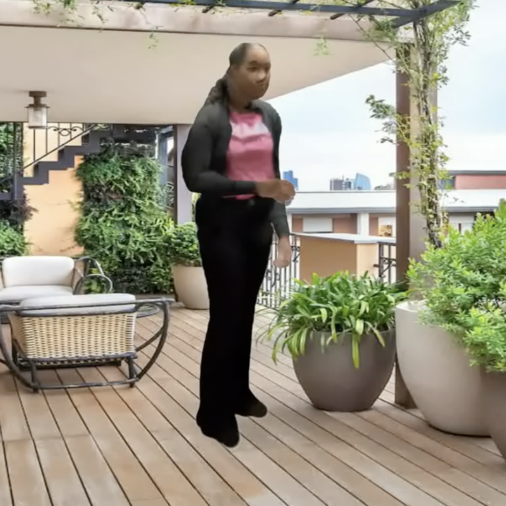} &
            \includegraphics[width=0.16\textwidth]{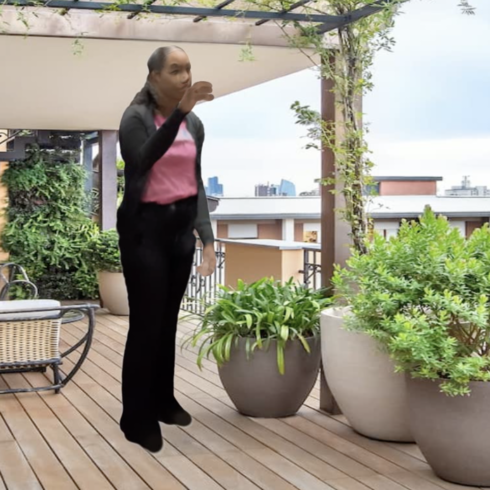} &
            \includegraphics[width=0.16\textwidth]{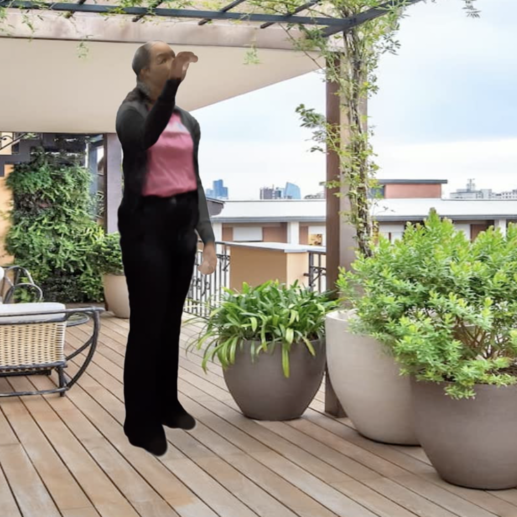} &
            \includegraphics[width=0.16\textwidth]{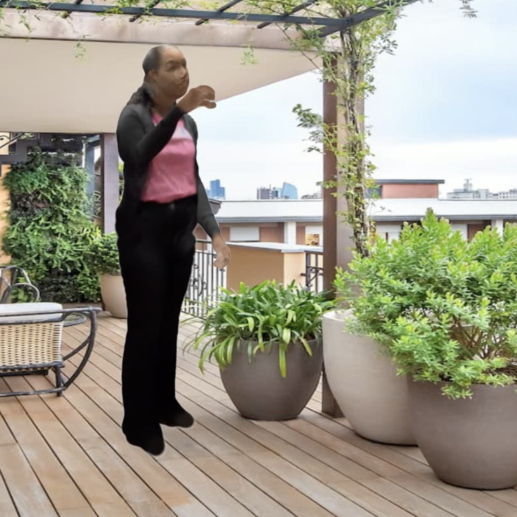} &
            \includegraphics[width=0.16\textwidth]{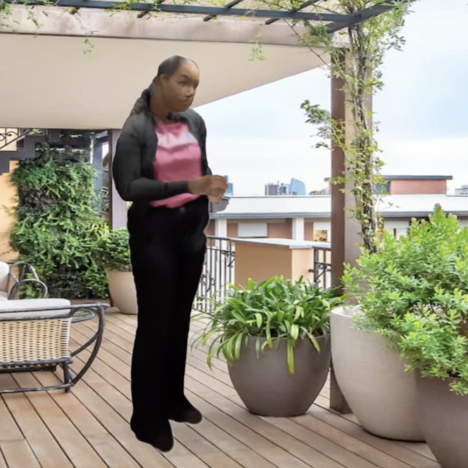} \\
        \end{tabular}
    }

    \caption{Examples of video frames at $t=\{0, 1, 2, 3, 4\}$ seconds from the source video (top), taken from NTU RGB+D dataset, and the target video (bottom), generated by our synthetic data generation method, where the pose alignment is consistent.}
    \label{fig:consistency_frame_grid_consistent}
\end{figure*}

\begin{figure*}[h]
    \centering
    \renewcommand{\arraystretch}{2.5}

    \begin{tabular}{@{}c@{\hskip 10pt}c@{\hskip 10pt}c@{\hskip 10pt}c@{\hskip 10pt}c@{}}
        \makebox[0.16\textwidth]{$t=0$} & 
        \makebox[0.16\textwidth]{$t=1$} & 
        \makebox[0.16\textwidth]{$t=2$} & 
        \makebox[0.16\textwidth]{$t=3$} & 
        \makebox[0.16\textwidth]{$t=4$} \\
    \end{tabular}

    \vspace{2pt}

    \fbox{
        \begin{tabular}{@{}c@{\hskip 10pt}c@{\hskip 10pt}c@{\hskip 10pt}c@{\hskip 10pt}c@{}}
        \includegraphics[width=0.16\textwidth]{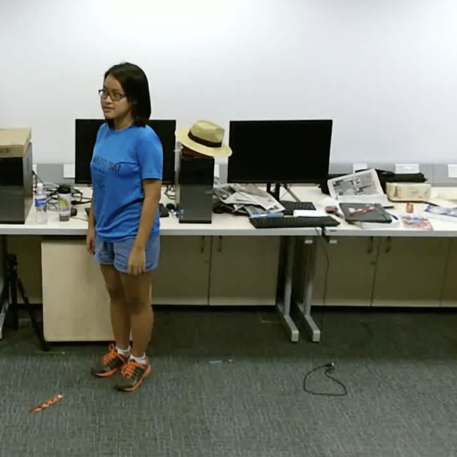} &
        \includegraphics[width=0.16\textwidth]{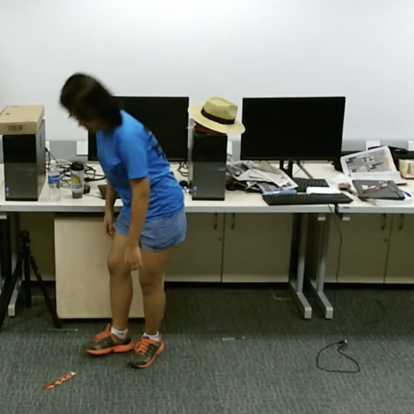} &
        \includegraphics[width=0.16\textwidth]{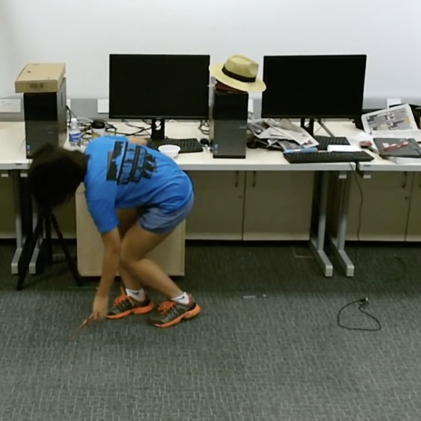} &
        \includegraphics[width=0.16\textwidth]{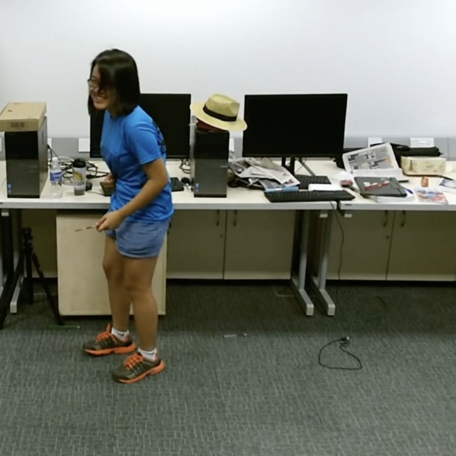} &
        \includegraphics[width=0.16\textwidth]{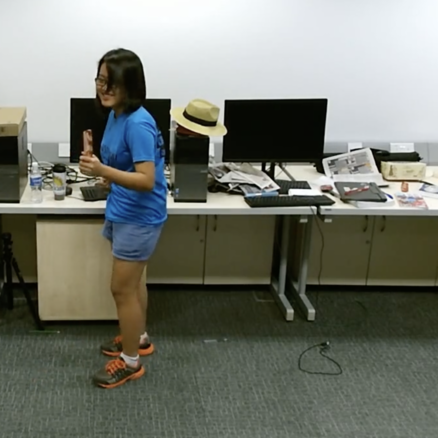} \\

        \includegraphics[width=0.16\textwidth]{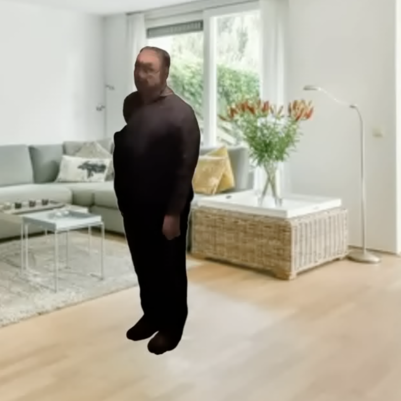} &
        \includegraphics[width=0.16\textwidth]{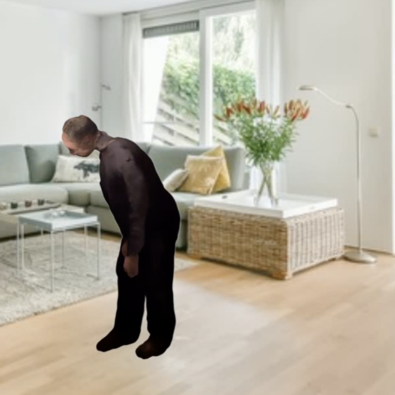} &
        \includegraphics[width=0.16\textwidth]{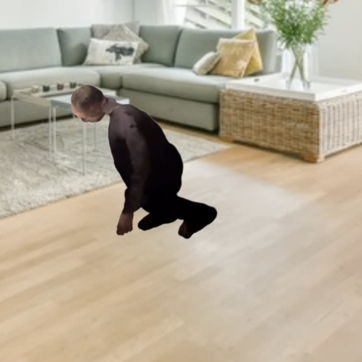} &
        \includegraphics[width=0.16\textwidth]{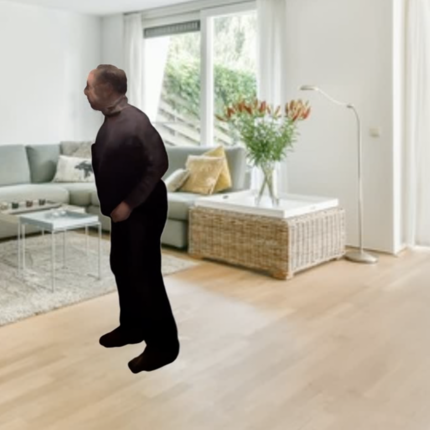} &
        \includegraphics[width=0.16\textwidth]{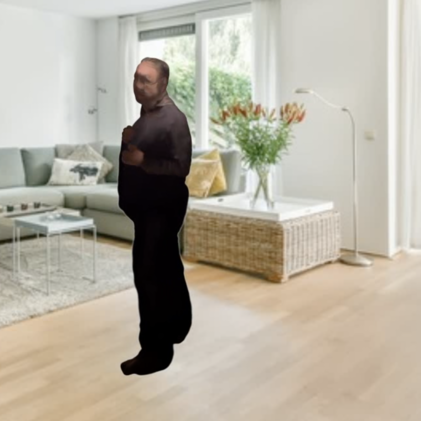} \\
        \end{tabular}
    }

    \vspace{0.3cm}

    \fbox{
        \begin{tabular}{@{}c@{\hskip 10pt}c@{\hskip 10pt}c@{\hskip 10pt}c@{\hskip 10pt}c@{}}
            \includegraphics[width=0.16\textwidth]{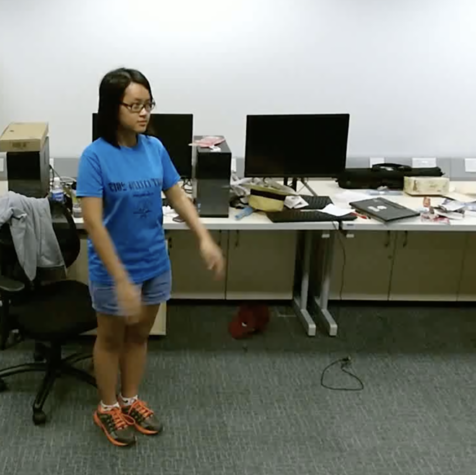} &
            \includegraphics[width=0.16\textwidth]{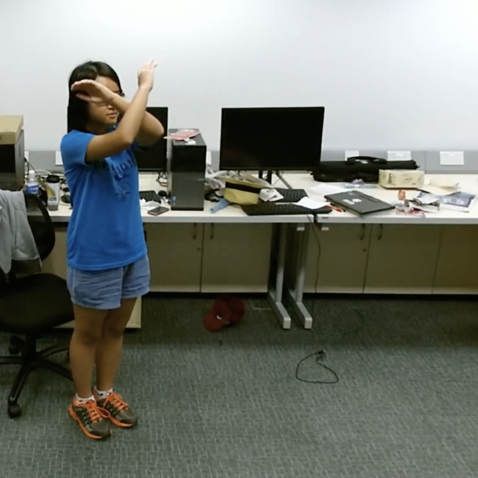} &
            \includegraphics[width=0.16\textwidth]{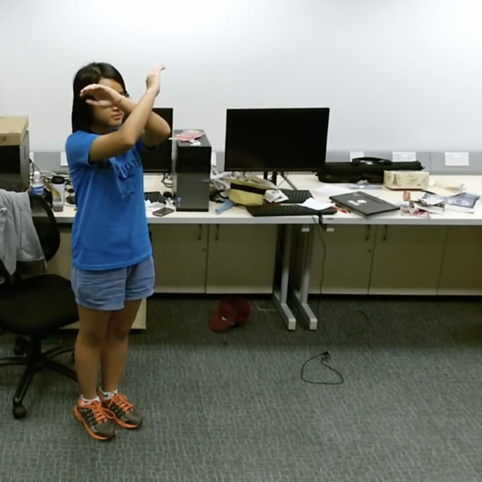} &
            \includegraphics[width=0.16\textwidth]{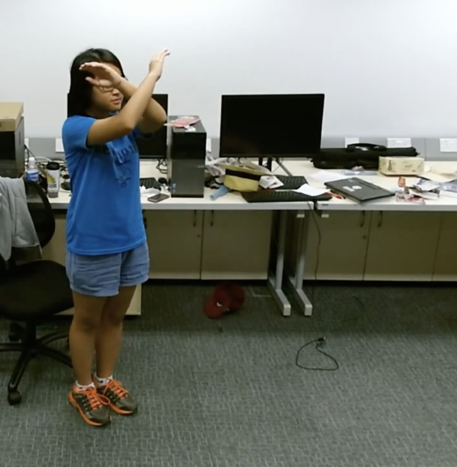} &
            \includegraphics[width=0.16\textwidth]{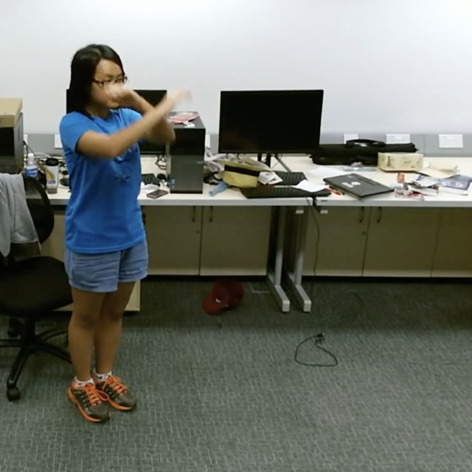} \\
    
            \includegraphics[width=0.16\textwidth]{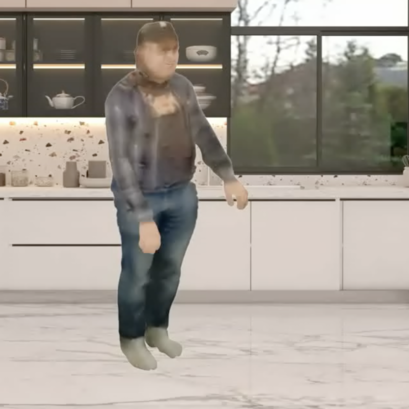} &
            \includegraphics[width=0.16\textwidth]{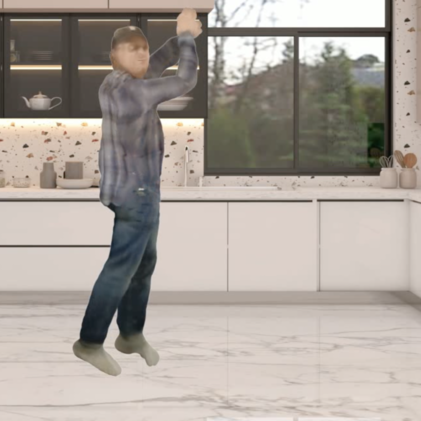} &
            \includegraphics[width=0.16\textwidth]{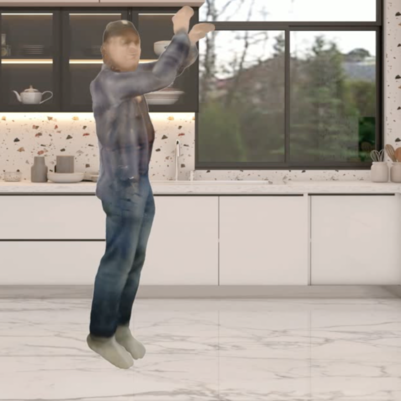} &
            \includegraphics[width=0.16\textwidth]{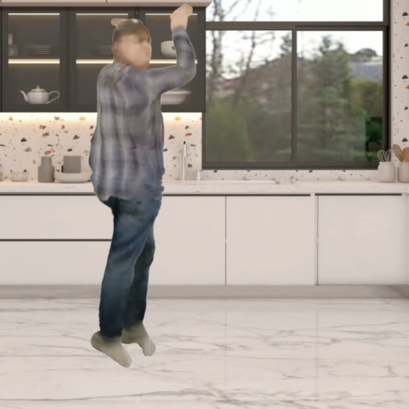} &
            \includegraphics[width=0.16\textwidth]{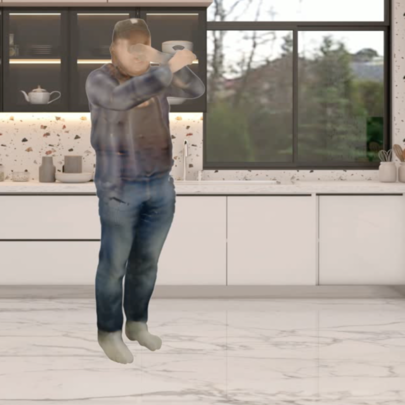} \\
        \end{tabular}
    }

    \vspace{0.3cm}

    \fbox{
        \begin{tabular}{@{}c@{\hskip 10pt}c@{\hskip 10pt}c@{\hskip 10pt}c@{\hskip 10pt}c@{}}
            \includegraphics[width=0.16\textwidth]{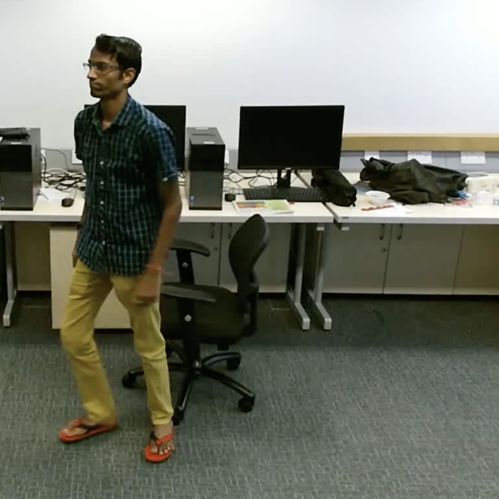} &
            \includegraphics[width=0.16\textwidth]{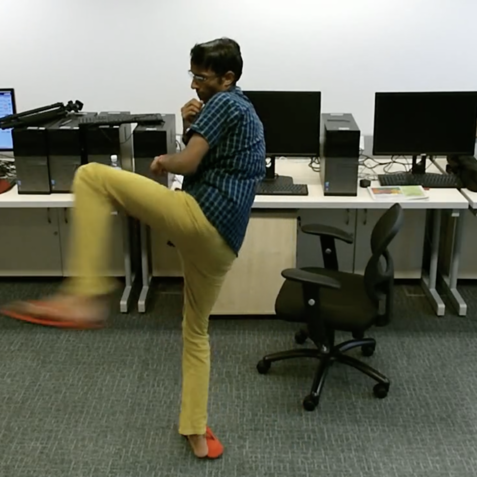} &
            \includegraphics[width=0.16\textwidth]{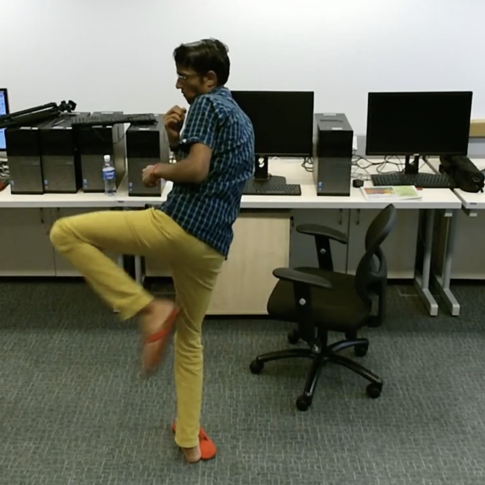} &
            \includegraphics[width=0.16\textwidth]{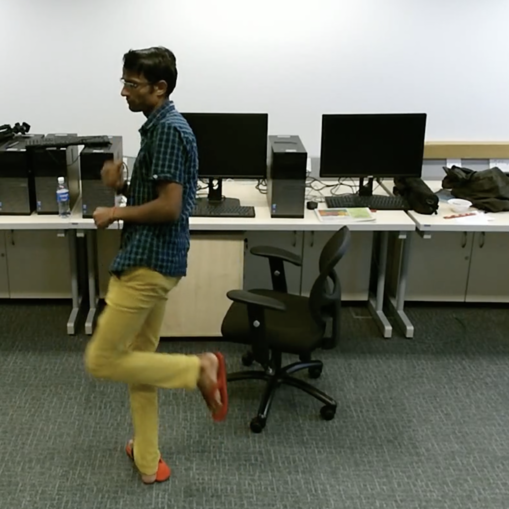} &
            \includegraphics[width=0.16\textwidth]{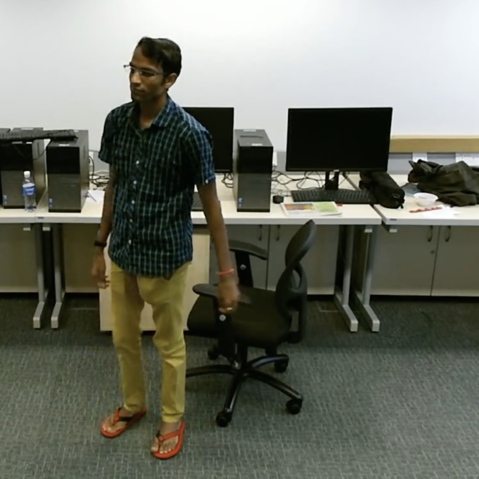} \\
    
            \includegraphics[width=0.16\textwidth]{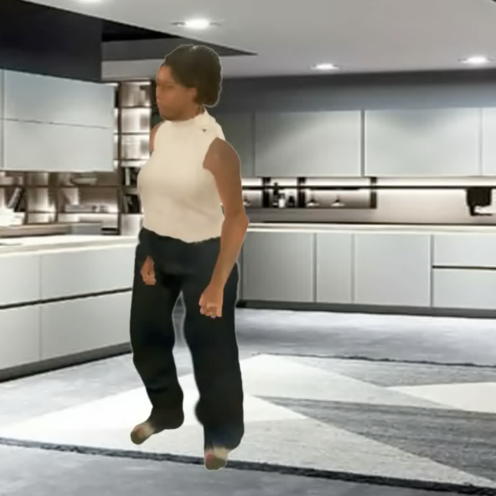} &
            \includegraphics[width=0.16\textwidth]{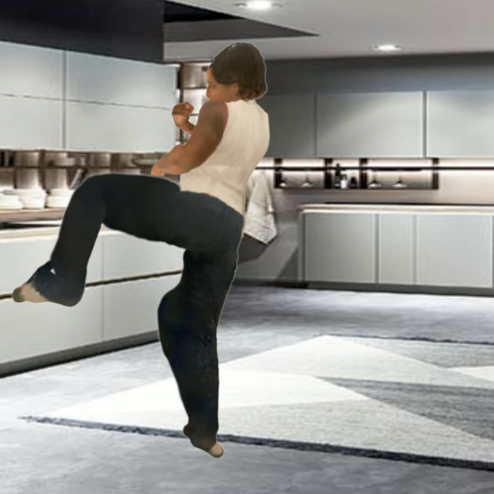} &
            \includegraphics[width=0.16\textwidth]{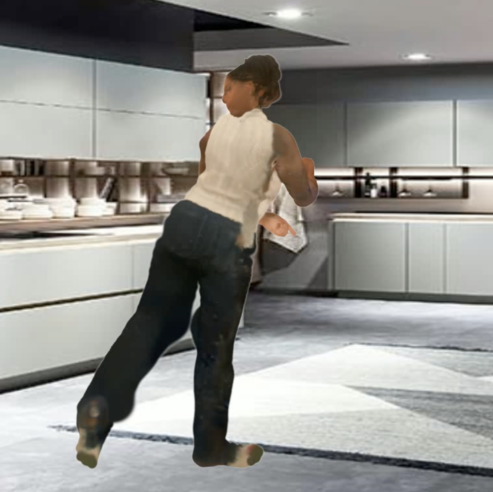} &
            \includegraphics[width=0.16\textwidth]{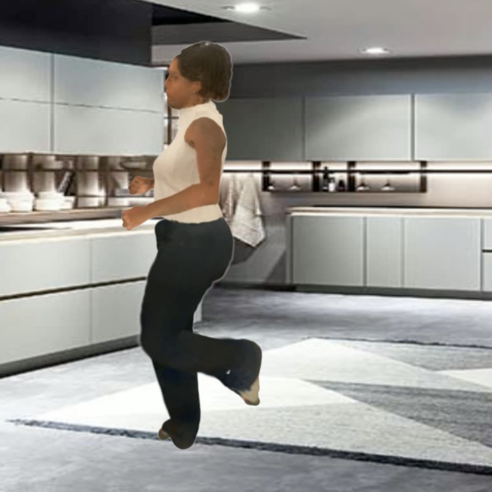} &
            \includegraphics[width=0.16\textwidth]{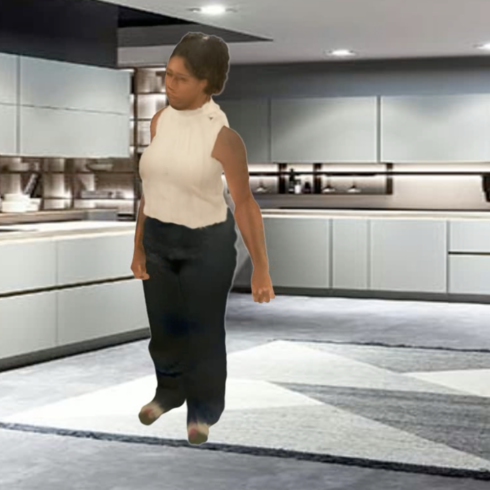} \\
        \end{tabular}
    }

    \caption{Examples of video frames at $t=\{0, 1, 2, 3, 4\}$ seconds from the source video (top), taken from NTU RGB+D dataset, and the target video (bottom), generated by our synthetic data generation method, where the pose alignment is inconsistent.}
    \label{fig:consistency_frame_grid_inconsistent}
\end{figure*}

\end{document}